\pdfoutput=1

\documentclass[letterpaper, 10 pt, conference]{ieeeconf}  % Comment this line out if you need a4paper

\IEEEoverridecommandlockouts                              % This command is only needed if 
\usepackage{cite}
\usepackage{amsmath,amssymb,amsfonts}
\usepackage{algorithmic}
\usepackage{graphicx}
\usepackage{textcomp}
\usepackage{xcolor}
\usepackage{lipsum}
\usepackage{bm}
\usepackage{float}
\DeclareTextSymbolDefault{\DH}{T1}

\usepackage{subfigure}

\graphicspath{ {./graphs/} }

\usepackage[utf8]{inputenc}
\usepackage{pgfplots}
\DeclareUnicodeCharacter{2212}{−}
\usepgfplotslibrary{groupplots,dateplot}
\usetikzlibrary{patterns,shapes.arrows}
\pgfplotsset{compat=newest}

\title{\LARGE \bf
A Computationally Efficient Learning-Based
Model Predictive Control for Multirotors under Aerodynamic Disturbances
}

\author{Babak Akbari and Melissa Greeff% <-this % stops a space
\thanks{The authors are with Robora Lab (www.roboralab.com), Queen's University; and affiliated with Ingenuity Labs Research Institute. E-mails: babak.akbari@queensu.ca, melissa.greeff@queensu.ca.}%
}

\begin{document}

\maketitle
\thispagestyle{empty}
\pagestyle{empty}

\begin{abstract}
Neglecting complex aerodynamic effects hinders high-speed yet high-precision multirotor autonomy. In this paper, we present a computationally efficient learning-based model predictive controller that simultaneously optimizes a trajectory that can be tracked within the physical limits (on thrust and orientation) of the multirotor system despite unknown aerodynamic forces and adapts the control input. To do this, we leverage the well-known differential flatness property of multirotors, which allows us to transform their nonlinear dynamics into a linear model. The main limitation of current flatness-based planning and control approaches is that they often neglect dynamic feasibility. This is because these constraints are nonlinear as a result of the mapping between the input, i.e., multirotor thrust, and the flat state. In our approach, we learn a novel representation of the drag forces by learning the mapping from the flat state to the multirotor thrust vector (in a world frame) as a Gaussian Process (GP). Our proposed approach leverages the properties of GPs to develop a convex optimal controller that can be iteratively solved as a second-order cone program (SOCP). In simulation experiments, our proposed approach outperforms related model predictive controllers that do not account for aerodynamic effects on trajectory feasibility, leading to a reduction of up to 55\% in absolute tracking error.
\end{abstract}

\section{INTRODUCTION}
% Multirotors
Multirotors are finding widespread use in applications such as infrastructure inspection, mapping operations \cite{c1}, payload delivery \cite{c2}, and search-and-rescue missions \cite{c3}. However, commercial autonomous multirotors typically operate at relatively low speeds. Control and planning algorithms must account for complex aerodynamic effects to enable advanced high-speed, high-precision applications. 
\begin{figure}
\centering
\label{fig:coordinates}
\includegraphics[width=0.48\textwidth, trim={0cm 0cm 0 0cm},clip]{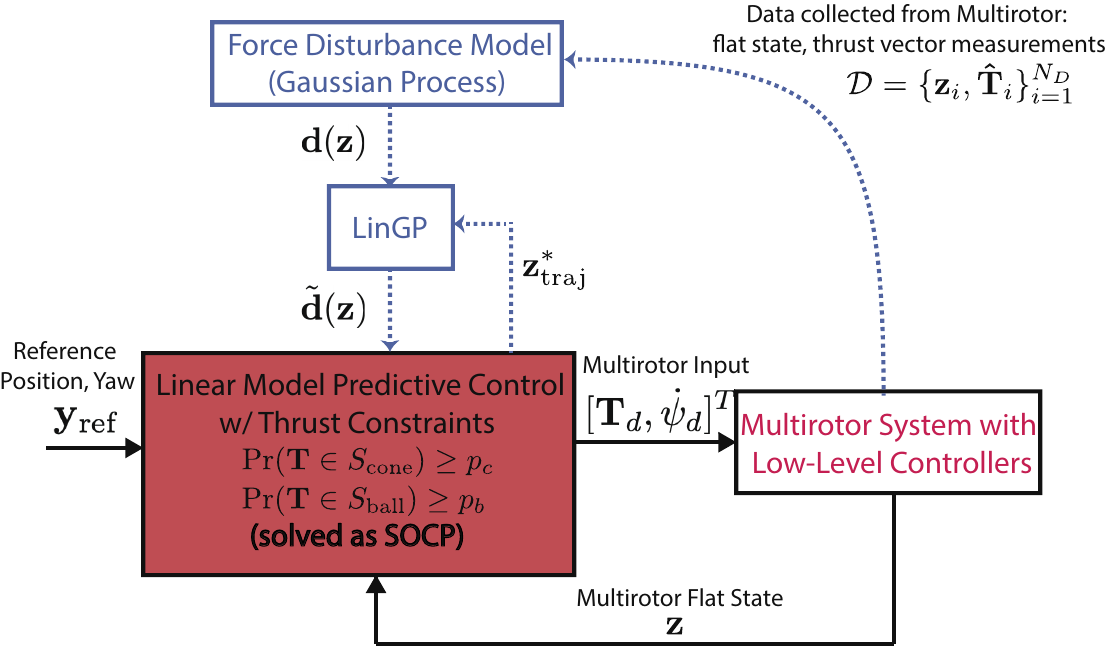}
\caption{Block diagram of our proposed learning-based MPC architecture. Our proposed approach: 1) learns the \textit{Force Disturbance Model} $\mathbf{d}(\mathbf{z})$,  as a Gaussian Process (GP), 2) linearizes the GP (\textit{LinGP}) about the current predicted optimal trajectory $\mathbf{z}^{*}_{\text{traj}}$ and 3) develops a linear MPC that enables high performance while ensuring dynamic feasibility (probabilistically) which is efficiently solved as a second-order cone program (SOCP).}
\vspace{-6mm}
\end{figure}
% Drag

Simple multirotor controllers have historically neglected aerodynamic effects, e.g., \cite{c4}, leading to significant tracking errors, especially at high speeds. One approach is to account for rotor drag, commonly modeled as linear in velocity \cite{c5}. However, this drag model simplification neglects the other complex drag effects, for example, the effect of induced and parasitic drag on the airframe and payload. Consequently, one common technique is to treat the aerodynamic effects as unknown disturbances and leverage disturbance rejection-based feedback controllers \cite{c6}. An alternative approach learns the drag effect through residual dynamics using a Gaussian Process (GP) \cite{c7} and uses it in a GP Model Predictive Controller (MPC). Similar GP MPC approaches have been applied to ground-based autonomous racing \cite{c8} and in offroad autonomy \cite{c9}. The main challenge with GP MPC is that the resulting optimal control problem is nonlinear and nonconvex, making it challenging to compute at high rates and sensitive to initial conditions.

% Differential Flatness
Neglecting drag effects, the common model of a multirotor is differentially flat with flat outputs consisting of its position and heading \cite{c10}. Similarly, the differential flatness property still holds for the model with linear rotor drag \cite{c5}. Leveraging differential flatness has become the standard approach for multirotor trajectory generation, e.g., \cite{c10}, \cite{c11}, \cite{c12}, etc. While this approach is both convex and efficient to solve, a significant drawback is that it does not guarantee that the optimized trajectory can be tracked under unmodeled aerodynamic effects within the multirotor’s physical limits. Differential flatness is also commonly used in feedback or feedforward linearization controllers \cite{c13}. These methods decouple an outer loop linear control, for example, linear MPC, and an inner loop feedback linearization technique, which incorporates the learned aerodynamic force, e.g., through a GP in \cite{c14} or \cite{c15}. In these approaches, the linear MPC, while efficiently solved as a quadratic program (QP), does not account for the effect of aerodynamic disturbances on input feasibility in the optimization. Consequently, it may plan infeasible trajectories to track given aerodynamic effects and input saturation constraints. One approach to address this is to introduce a tracking cost that measures the ability of inner loop controllers to follow the optimized reference trajectory \cite{c16}. 

% Problem 
In this paper, we address the problem of accounting for aerodynamic forces through a lightweight implementation that can be computed in real-time while operating within the physical limits of our multirotor platform (e.g., maximum thrust produced by rotors). Our proposed drag-aware model predictive controller simultaneously adapts the desired trajectory and the control input. Similar to efficient planning and optimal control algorithms for multirotors \cite{c10} \cite{c13}, we leverage the property of differential flatness but take into account saturation limits, i.e., the maximum thrust produced by the rotors and maximum pitch and roll, without losing the convexity of the optimization problem. Similar to \cite{c7}, we model the aerodynamic forces as GPs. However, we learn an alternative mapping from the flat state (as opposed to state) to the thrust vector (as opposed to the collective thrust and body torques). By doing this and leveraging the properties of GPs, we show how we can optimize a feasible trajectory, taking into account learned aerodynamic forces, with high probability as a second-order cone program (SOCP). We simultaneously leverage the same GPs to adapt the control input for the expected aerodynamic force to track this optimized feasible trajectory. The contributions of this paper are three-fold:
\begin{itemize}
    \item We learn a novel representation of the drag forces by learning the mapping from the multirotor flat state to the thrust vector (represented in an inertial world frame). In this paper, we will leverage a GP to learn this mapping.
    \item We develop a novel model predictive controller that leverages differential flatness and our learned aerodynamic GP model to optimize a feasible trajectory as a second-order cone program (SOCP). The SOCP structure of this convex optimization allows us to solve it efficiently in real time.
    \item We demonstrate in simulation how our proposed approach outperforms related model predictive control approaches that do not account for aerodynamic effects on trajectory feasibility.  
\end{itemize}
\section{PROBLEM STATEMENT}
\noindent
As illustrated in Fig. \ref{fig:coordinates}, we use an inertial world frame $W$ and a body frame $B$ fixed at the multirotor's center of mass. In this paper, we consider the nonlinear multirotor dynamics $ \dot{   {\mathbf{x}}} = f(\mathbf{   {x}},\mathbf{   {u}})$ as
\begin{equation}
    \begin{aligned}
        \dot{\mathbf{p}} &= \mathbf{v},\\
        \mathbf{a} &= \dot{\mathbf{v}} = -g\mathbf{z}_W + \frac{T}{m}\mathbf{z}_{B} + \mathbf{F}_{d}(\mathbf{{x}}),\\
        \dot{\mathbf{R}} &= \mathbf{R}\hat{\bm{\omega}},\\
        \dot{\bm{\omega}} &= \mathbf{J}^{-1}(\bm{\tau} - \bm{\omega} \times \bm{J} \bm{\omega}),
        \label{eq_uav_system}
    \end{aligned}
\end{equation}
where the state $\mathbf{   {x}} = [\mathbf{p}, \mathbf{v}, \mathbf{R}, \bm{\omega}]^T$  comprises of its position $\mathbf{p}$ and velocity $\mathbf{v}$ in the world frame, the orientation $\mathbf{R} = [\mathbf{x}_B, \mathbf{y}_B, \mathbf{z}_B]$ of the body frame $B$ with respect to $W$ and body rates $\bm{\omega}$. The input $   {\mathbf{u}} = [T, \bm{\tau}]^T$ comprises of the collective thrust $T \in \mathbb{R}$ and body torques $\bm{\tau} \in \mathbb{R}^3$. In our model (\ref{eq_uav_system}), $\mathbf{F}_{d}(\mathbf{   {x}})$ is an \textit{unknown} aerodynamic disturbance, for example, as a result of drag or wind. 

Neglecting this disturbance $\mathbf{F}_{d}(\mathbf{   {x}})$, it is well known that the multirotor model (\ref{eq_uav_system}) is 
\textit{differentially flat} in output $\mathbf{y} = [\mathbf{p}, \psi]^T$ comprising of position $\mathbf{p}$ and yaw $\psi$, see \cite{c10}. Specifically, the differential flatness property of (\ref{eq_uav_system}) allows us to transform the nonlinear dynamics into an equivalent linear system:
\begin{equation}
    \dot{   {\mathbf{z}}} = \mathbf{A}\mathbf{   {z}} + \mathbf{B}   {\mathbf{v}},
    \label{eq_lin_flat}
\end{equation}
where \(\mathbf{   {z}} = [\mathbf{p}, \mathbf{v}, \mathbf{a}, \mathbf{j}, \mathbf{\psi}]^T\) is the flat state comprising of the multirotor position $\mathbf{p}$, velocity $\mathbf{v}$, acceleration $\mathbf{a}$, jerk $\mathbf{j}$ in the world frame $W$ and yaw $\mathbf{\psi}$. The flat input $   {\mathbf{v}} = [\mathbf{s}, \dot{\psi}]^T$ comprises of the snap $\mathbf{s}$ in the world frame and yaw rate $\dot{\psi}$.

\paragraph*{Assumption 1} $\mathbf{F}_{d}(\mathbf{   {x}})$ belongs to the class of functions such that the system (\ref{eq_uav_system}) retains the differential flatness property in the output $\mathbf{y} = [\mathbf{p}, \psi]^T$. 

\paragraph*{Remark} While accounting for linear rotor drag in (\ref{eq_uav_system}) retains the differential flatness, see \cite{c5}, this is still a significant assumption of our work. We leave defining this class of functions for which differential flatness in $\mathbf{y} = [\mathbf{p}, \psi]^T$ is retained for future work.  

\noindent
Multirotor trajectory generation commonly leverages differential flatness where sufficiently smooth trajectories in $\mathbf{y}$ can be be tracked by the system. However, these approaches currently neglect dynamic feasibility, e.g., \cite{c10} or \cite{c11}, or implement conservative box constraints on the flat states, see \cite{c13}. Our objective, given Assumption 1, is to optimize a collision-free trajectory that can be tracked in real-time despite unknown aerodynamic disturbances $\mathbf{F}_{d}(\mathbf{   {x}})$ and is dynamically feasible with high probability. To generate a dynamically feasible trajectory, the following constraints need to be satisfied:
\begin{equation}
    \begin{aligned}
    0 \leq T & \leq T_{max}, \\
    |\theta| & \leq \theta_{max}, \\
    |\phi| & \leq \phi_{max},
    \label{eq_dyn_feasibility}
    \end{aligned}
\end{equation}
where $T_{max}$ is the maximum collective thrust that the rotors generate, $\theta$ and $\phi$ are the pitch and roll angles respectively, and $\theta_{max}$ and $\phi_{max}$ are the selected (often for safety) maximum pitch and roll angles respectively. As is common, we assume  $\theta_{max} = \phi_{max}$.
\begin{figure}
\vspace{2mm}
\centering
\includegraphics[width=0.4\textwidth]{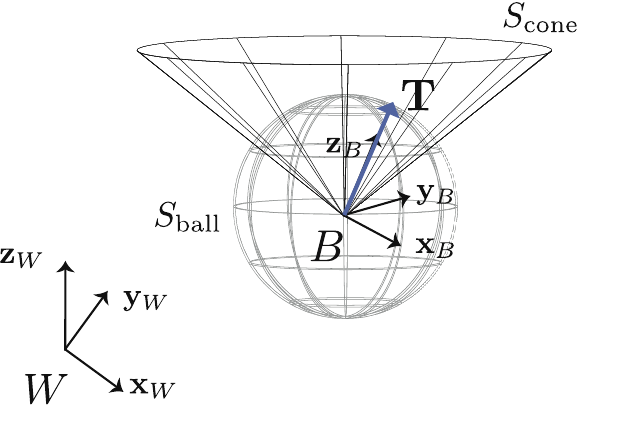}
\caption{Schematic of multirotor coordinate frames and the proposed constraints on thrust vector $\mathbf{T}$ (blue) in (\ref{eq_ball}) and (\ref{eq_cone}).}
\label{fig:coordinates}
\vspace{-5mm}
\end{figure}

\section{BACKGROUND}
%\subsection{Multirotor UAV dynamics}
%We consider the dynamical model of a quadrotor. According to this
%model, the dynamics of the position $\mathbf{p}$, velocity $\mathbf{v}$, orientation
%$R = [\mathbf{x}_B, \mathbf{y}_B, \mathbf{z}_B]$, and body rates $\bm{\omega}$ can be written as:
%\begin{equation}
%    \begin{aligned}
%        \dot{\mathbf{p}} &= \mathbf{v}\\
%        \dot{\mathbf{v}} &= -g\mathbf{z}_W + c\mathbf{z}_{B} - \mathbf{R}\mathbf{D}\mathbf{R}^T\mathbf{v}\\
%        \dot{\mathbf{R}} &= \mathbf{R}\hat{\bm{\omega}}\\
 %       \dot{\bm{\omega}} &= \mathbf{J}^{-1}(\bm{\tau} - \bm{\omega} \times \bm{J} \bm{\omega} - \bm{\tau}_g - \mathbf{A}\mathbf{R}^{T}\mathbf{v} - \mathbf{B}\bm{\omega})\\
%    \end{aligned}
%\end{equation}
%where $c$ is the mass-normalized collective thrust,
%$\mathbf{D} = diag (dx, dy, dz)$ is a constant diagonal matrix formed
%by the mass-normalized rotor-drag coefficients, $\hat{\bm{\omega}}$ is a
%skew-symmetric matrix formed from $\bm{\omega}$, $\mathbf{J}$ is the quadrotor’s
%inertia matrix, $\bm{\tau}$ is the three dimensional torque input, $\bm{\tau}_g$
%are gyroscopic torques from the propellers, and $\mathbf{A}$ and $\mathbf{B}$
%are constant matrices.

\subsection{Flatness-Based MPC}
In MPC, we approximate the continuous-time system (\ref{eq_lin_flat}) with the discrete-time system
$\mathbf{z}_{i+1} = \mathbf{A}_d \mathbf{z}_i + \mathbf{B}_d \mathbf{v}_i$ where $\mathbf{z}_i$ is the flat state $\mathbf{z}$ at time step $i$ with $\delta$ as the discretization time interval used for $\mathbf{A}_d$ and $\mathbf{B}_d$. Flatness-Based MPC takes the current flat state $\mathbf{z}_{i} = \mathbf{z}_{\text{init}}$ at each time step $i$ and produces a sequence of optimal flat states $\mathbf{z}^{*}_{0:N}$ and control commands $\mathbf{v}^{*}_{0:N-1}$ where the notation $^{*}$ denotes the optimal solution and  $_{0:N}$ denotes the value for each time step from the current time step $i$ to $i+N$ where $N \in \mathbb{Z}$ is the prediction horizon. It does this by solving an optimization problem online, using a multiple shooting scheme, see \cite{c19}, and then applying the first control command, after which the optimization problem is solved again in the next state. Specifically, we solve the optimal control problem (OCP) using the discretized linear dynamics (\ref{eq_lin_flat}) as:
\begin{equation}
\begin{aligned}
	\min_{\mathbf{z}_{0:N}, \mathbf{v}_{0:N-1}} & \quad J(\mathbf{z}_{0:N}, \mathbf{v}_{0:N-1}, \mathbf{y}^{ref}_{0:N}) \\
\textrm{s.t.} \quad & \mathbf{z}_{k+1} = \mathbf{A}_d \mathbf{z}_k + \mathbf{B}_d \mathbf{v}_k \quad \forall k = 0,..., N-1\\
  &\mathbf{z}_{k} \in \mathcal{Z} \quad \forall k = 0,..., N-1   \\
  &\mathbf{v}_{k} \in \mathcal{V} \quad  \forall k = 0,..., N-1  \\
  &\mathbf{z}_{0} = \mathbf{z}_{\text{init}}
  \label{eq_fmpc}
\end{aligned}
\end{equation}
and $J(\cdot)$ is generally selected as a quadratic cost:
\vspace{-2mm}
\begin{equation}
    J(\cdot) = \sum_{k=1}^{N} (\mathbf{C} \mathbf{{z}}_k - \mathbf{y}^{ref}_k)^T\mathbf{Q}(\mathbf{C} \mathbf{{z}}_k - \mathbf{y}^{ref}_k) + \mathbf{v}_{k-1}^T \mathbf{R} \mathbf{v}_{k-1},
    \label{eq_cost}
\end{equation}
where $\mathbf{y}_k = \mathbf{C} \mathbf{z}_k$ and $\mathbf{Q}$ and $\mathbf{R}$ are positive definite matrices that weight the position error and control effort respectively. The state $\mathcal{Z}$ and input $\mathcal{V}$ constraint sets are often selected (conservatively) as linear constraints such that the optimization (\ref{eq_fmpc}) is not only convex but can be solved as a Quadratic Program (QP). The incorrect selection of these constraints in (\ref{eq_fmpc}) may lead to violations of the dynamic feasibility constraints in (\ref{eq_dyn_feasibility}). A conservative selection will lead to sub optimal performance. Furthermore, given that the aerodynamic disturbance $\mathbf{F}_{d}(\mathbf{x})$ is unknown prior to flight, it may be challenging to select the constraint sets $\mathcal{Z}$ and $\mathcal{V}$ (even conservatively) a priori. 

\subsection{Gaussian Processes (GPs)}
GP regression is a nonparametric method that can be used to approximate a nonlinear function, \(\mathbf{d}(\mathbf{z}): \mathbb{R}^{dim(\mathbf{z})} \rightarrow \mathbb{R}\) from input $\mathbf{z}$ to function values $\mathbf{d}(\mathbf{z})$. We use the notation $\mathbf{d}(\mathbf{z})$ to be consistent with Section IV. Each function value $\mathbf{d}(\mathbf{z})$, evaluated at input $\mathbf{z}$, is a random variable, and any finite number of these random variables have a joint Gaussian distribution. GP regression requires a prior mean for $\mathbf{d}(\mathbf{z})$, which is generally set to zero, and a kernel function $k(., .)$ to associate a similar level of uncertainty to inputs close to each other. For example, a common kernel function is the squared-exponential (SE) function:
\vspace{-2mm}
\begin{equation}
       k(\mathbf{z},\mathbf{z}') = \sigma_{\eta}^2\exp{(-\frac{1}{2}(\mathbf{z}-\mathbf{z}')^T\mathbf{M}^{-2}(\mathbf{z} -\mathbf{z}'))}+\delta_{ij}\sigma_{\omega}^2\\
       \label{eq_se_kernel}
\end{equation}
which is characterized by three types of hyperparameters: the prior variance $\sigma^2_{\eta}$, measurement noise $\sigma^2_{\omega}$ where $\delta_{ij}=1$
if $i = j$ and 0 otherwise, and the length scales, or the diagonal elements of the diagonal matrix $\mathbf{M}$, which encode a measure of how quickly the function $\mathbf{d}(\mathbf{z})$ changes with respect to
$\mathbf{z}$.

\section{METHODOLOGY}

We enforce constraints (\ref{eq_dyn_feasibility}) by imposing equivalent constraints on the thrust vector $\mathbf{T} = [T_x, T_y, T_z]^T = T \mathbf{z}_B \in \mathbb{R}^3$ in world frame $W$, see Fig. 2. Specifically, we can enforce (\ref{eq_dyn_feasibility}), without  approximation, by ensuring that $\mathbf{T}$ remains in two sets $S_{\text{ball}}$ and $S_{\text{cone}}$ where:
\begin{equation*}
    S_{\text{ball}} = \{\mathbf{T} \quad | \quad ||\mathbf{T}||_2 \leq T_{max} \},  
\end{equation*}
\begin{equation*}
    S_{\text{cone}} = \{\mathbf{T} \quad | \quad ||[T_x, T_y]^T||_2 \leq \tan{\theta_{max}} T_z \},
\end{equation*}
and the notation $||\cdot||_2$ denotes the two norm. In this work, we consider that aerodynamic force $\mathbf{F}_d(\mathbf{x})$ may be \textit{unknown}, and an upper bound is not given. We, therefore, aim to enforce feasibility constraints probabilistically as:
\begin{equation}
    \text{Pr}(\mathbf{T} \in S_{\text{ball}}) \geq p_b,
     \label{eq_ball}
\end{equation}
\begin{equation}
    \text{Pr}(\mathbf{T} \in S_{\text{cone}}) \geq p_c,
    \label{eq_cone}
\end{equation}
where $p_b$ and $p_c$ are user-selected probabilities. In our proposed approach, we solve a linear model predictive control (\ref{eq_fmpc}) subject to (\ref{eq_ball}) and (\ref{eq_cone}) as a SOCP by (i) Learning the force disturbance as a GP; (ii) Linearizing the GP (through a method LinGP \cite{c18}) about the current optimal trajectory $\mathbf{z}^*_{traj}$; (iii) Describing the stochastic constraints (\ref{eq_ball}) and (\ref{eq_cone}) as second-order cone constraints on the optimization variables $\mathbf{z}_{0:N}$ and $\mathbf{v}_{0:N-1}$.

\subsection{Learning Force Disturbance as a Gaussian Process}
From (\ref{eq_uav_system}), we can describe the relationship between the thrust vector and the state as $\mathbf{T} = T \mathbf{z}_B = m\mathbf{a} + m g \mathbf{z}_W - \mathbf{F}_d(\mathbf{x})$ or plugging in the transformation $\mathbf{x} = \gamma(\mathbf{z})$ between state $\mathbf{x}$ and flat state $\mathbf{z}$ (from differential flatness -- see Assumption 1),
\begin{equation}
    \mathbf{T} = m\mathbf{a} + m g \mathbf{z}_W - \mathbf{d}(\mathbf{z}),
    \label{eq_thrust}
\end{equation}
where $\mathbf{d}(\mathbf{z}) = \mathbf{F}_d(\gamma(\mathbf{z})) = [d_x(\mathbf{z}), d_y(\mathbf{z}), d_z(\mathbf{z})]^T$ and  acceleration $\mathbf{a} = [\mathbf{a}_x, \mathbf{a}_y, \mathbf{a}_z]^T$. 
\paragraph*{Assumption 2} We assume that  $d_x(\mathbf{z}), d_y(\mathbf{z}), d_z(\mathbf{z})$ are independent functions. 

We learn $d_x(\mathbf{z}), d_y(\mathbf{z}), d_z(\mathbf{z})$ as independent GPs. For example, we use our GP to predict the function value $d_x(\mathbf{z})$ at any query point $\mathbf{z}^*$ based on $N_D$ noisy observations, $\mathcal{D} = \{ \mathbf{z}_i,  \hat{d}_x \}^{N_D}_{i=1}$ where $\hat{d}_x = m \mathbf{a}_x - \hat{T}_x(\mathbf{z}_i)$ is a measurement of $d_x(\mathbf{z})$ with zero mean Gaussian noise $\sigma_\omega^2$. The predicted mean and variance at the query point $\mathbf{z}^*$ conditioned on the observed data $\mathcal{D}$ are \cite{c17}:
%\vspace{-3mm}
\begin{equation}
	\mu_x(\mathbf{z}^*) = \mathbf{k}(\mathbf{z}^*, \mathbf{Z}) \mathbf{K}^{-1} \mathbf{\hat{D}}_{x}, 
	 \label{eq_mean}
\end{equation}
	 \vspace{-4mm}
\begin{equation}
	\sigma_{x}^2(\mathbf{z}^*) = k(\mathbf{z}^*, \mathbf{z}^*) - \mathbf{k}(\mathbf{z}^*, \mathbf{Z}) \mathbf{K}^{-1} \mathbf{k}^T(\mathbf{z}^*, \mathbf{Z}),
	\label{eq_var}
\end{equation}
\noindent
where $\mathbf{\hat{D}}_{x} =[\hat{d}_{x}(\mathbf{z}_1), ..., \hat{d}_{x}(\mathbf{z}_{N_D})]^T$ is the vector of observed function values, the covariance matrix has entries $\mathbf{K}_{(i,j)} = k(\mathbf{z}_i, \mathbf{z}_j), \quad i,j \in {1, ..., N_D}$, and $\mathbf{k}(\mathbf{z}^*, \mathbf{Z}) = [k(\mathbf{z}^*, \mathbf{z}_1), ..., k(\mathbf{z}^*, \mathbf{z}_{N_D})]$ is the vector of the covariances between the query point $\mathbf{z}^*$ and the observed data points in $\mathcal{D}$. Consequently, we can predict $T_x(\mathbf{z}^*)|_{\mathbf{D}_x} \sim \mathcal{N}(\mu_{T,x}(\mathbf{z}^*), \sigma^2_{T,x}(\mathbf{z}^*))$ where $\mu_{T,x}(\mathbf{z}^*) = m \mathbf{a}_x - \mu_x(\mathbf{z}^*)$ and $\sigma^2_{T,x}(\mathbf{z}^*) = \sigma_{x}^2(\mathbf{z}^*)$. Similarly, using the same approach for the $y$ and $z$ thrust components we can infer $T_y(\mathbf{z}^*)|_{\mathbf{D}_y} \sim \mathcal{N}(\mu_{T,y}(\mathbf{z}^*), \sigma^2_{T,y}(\mathbf{z}^*))$ and $T_z(\mathbf{z})|_{\mathbf{D}_z} \sim \mathcal{N}(\mu_{T,z}(\mathbf{z}^*), \sigma^2_{T,z}(\mathbf{z}^*))$.

\subsection{Linearizing Gaussian Process}

Given an input vector $\mathbf{z}^*$, we will assume that $d_x$ is differentiable at $\mathbf{z}^*$. We also assume that the covariance function $k(\cdot,\cdot)$ is differentiable, which is typically the case, e.g., see \cite{c18}. Recall that the GP of $d_x$ is essentially a (posterior) distribution over the space of realizations of $d_x$. We linearize $d_x$ around $\mathbf{z}^*$ using a method LinGP, see \cite{c18}, as $d_x(\mathbf{z}^* + \Delta_{\mathbf{z}}) \approx d_x(\mathbf{z}^*) + \Delta_{\mathbf{z}}^T \nabla_{\mathbf{z}}d_x(\mathbf{z}^*)
    \label{eq_lin_d} $, where $\nabla_{\mathbf{z}}d_x(\mathbf{z}^*)$ is the gradient of $d_x$ at $\mathbf{z}^*$ and $\Delta_{\mathbf{z}}$ is the displacement from $\mathbf{z}^*$. We can rewrite this as:
\begin{equation}
    d_x(\mathbf{z}^* + \Delta_{\mathbf{z}}) \approx \mathbf{\bar{z}}^T \bar{d}_x(\mathbf{z}^*) := \tilde{d}_x(\Delta_{\mathbf{z}}),
    \label{eq_lin_d_2}
\end{equation}
where we use augmented vector 
\begin{equation}
    \mathbf{\bar{z}} = [1, \Delta_{\mathbf{z}}]^T,
    \label{eq_aug}
\end{equation}
and $\bar{d}_x(\mathbf{z}^*) = [d_x(\mathbf{z}^*),  \nabla_{\mathbf{z}}^T d_x(\mathbf{z}^*)]^T$. As noted in \cite{c18}, $\bar{d}_x(\mathbf{z}^*)$ is a random vector of length $(n+1)$, where $\mathbf{z}^* \in \mathbb{R}^n$, which, through the inner product with the vector $\mathbf{\bar{z}}$, results in the random variable $\tilde{d}_x(\Delta_{\mathbf{z}})$. We will approximate the original GP of $d_x$ around $\mathbf{z}^*$ by a GP of the linearized function $\tilde{d}_x(\Delta_{\mathbf{z}})$ in (\ref{eq_lin_d_2}) which we will call \textit{LinGP} in Fig. 1.

Given that differentiation is a linear operator, $\bar{d}_x(\mathbf{z}^*)$ is also a GP (multivariate) that can be derived from the original GP model of $d_x(\mathbf{z}^*)$. In other words, $\bar{d}_x(\mathbf{z}^*)|{\mathcal{D}}$ defines a posterior distribution, derived from the posterior $d_x(\mathbf{z}^*)|{\mathcal{D}}$ and its gradient at $\mathbf{z}^*$. Specifically, the posterior distribution of of $\bar{d}_x(\mathbf{z}^*)$ conditioned on the data $\mathcal{D}$ is $\bar{d}_x(\mathbf{z}^*)|{\mathcal{D}} \sim \mathcal{N}(\bar{\mu}_x(\mathbf{z}^*), \bar{V}_x(\mathbf{z}^*))$ with mean:
        \begin{equation}
    \bar{\mu}_x(\mathbf{z}^*) = 
    \begin{bmatrix}
        \mathbf{k}(\mathbf{z}^*, \mathbf{Z}) \\ K^{(1,0)}(\mathbf{z}^*, \mathbf{Z})
    \end{bmatrix} \mathbf{K}^{-1} \mathbf{\hat{D}}_{x}, 
\end{equation}
where the notation $K^{(1,0)}(\mathbf{z}^*, \mathbf{Z}) = [K^{(1,0)}(\mathbf{z}^*, \mathbf{z}_1), ..., K^{(1,0)}(\mathbf{z}^*, \mathbf{z}_{N_D})]$ and $K^{(1,0)}(\mathbf{z}, \mathbf{z}') = \nabla_{\mathbf{z}} k(\mathbf{z}, \mathbf{z}')$ is used to denote the gradient of the kernel, for example (\ref{eq_se_kernel}), with respect to its first argument. We use the notation $K^{(0,1)}(\mathbf{z}, \mathbf{z}') = D_{\mathbf{z}'} k(\mathbf{z}, \mathbf{z}')$ to denote the Jacobian of the kernel, for example (\ref{eq_se_kernel}), with respect to its second argument. Similarly, $K^{(0,1)}(\mathbf{Z}, \mathbf{z}^*) = [K^{(0,1)}(\mathbf{z}_1, \mathbf{z}^*), ..., K^{(0,1)}(\mathbf{z}_{N_D}, \mathbf{z}^*)]$. The notation $K^{(1,1)}(\mathbf{z}, \mathbf{z}') \in \mathbb{R}^{n \times n}$ denotes a matrix with entries $i,j$ as $K^{(1,1)}_{(i,j)}(\mathbf{z}, \mathbf{z}') = \frac{\partial^2}{\partial \mathbf{z}_i \partial \mathbf{z}_j} k(\mathbf{z}, \mathbf{z}')$. The covariance matrix $\bar{V}_x(\mathbf{z}^*)$ is then given by:
\begin{multline}
        \bar{V}_x(\mathbf{z}^*) = \begin{bmatrix}
    k(\mathbf{z}^*, \mathbf{z}^*) & K^{(0,1)}(\mathbf{z}^*, \mathbf{z}^*) \\
    K^{(1,0)}(\mathbf{z}^*, \mathbf{z}^*) & K^{(1,1)}(\mathbf{z}^*, \mathbf{z}^*)
    \end{bmatrix} - \\
    \begin{bmatrix}
        \mathbf{k}(\mathbf{z^*, \mathbf{Z}}) \\ K^{(1,0)}(\mathbf{z}^*, \mathbf{Z})
    \end{bmatrix} \mathbf{K}^{-1} \begin{bmatrix}
        \mathbf{k}(\mathbf{z}^*, \mathbf{Z})^T & K^{(0,1)}(\mathbf{Z}, \mathbf{z}^*),
    \end{bmatrix}
    \label{eq_cov_matrix}
\end{multline}
where $\bar{V}_x(\mathbf{z}^*) \in \mathbb{R}^{(n+1) \times (n+1)}$ is positive semi-definite. Using (\ref{eq_lin_d_2}), it follows that $\tilde{d}_x(\Delta_{\mathbf{z}})|_{\mathcal{D}} \sim \mathcal{N}(\tilde{\mu}_x(\Delta_{\mathbf{z}}), \tilde{\sigma}^2_x(\Delta_{\mathbf{z}}))$ where:
\begin{equation}
    \tilde{\mu}_x(\Delta_{\mathbf{z}}) = \bar{\mu}_x(\mathbf{z}^*)^T \mathbf{\bar{z}},
    \label{eq_lingp_mean}
\end{equation}
\begin{equation}
    \tilde{\sigma}^2_x(\Delta_{\mathbf{z}}) = \mathbf{\bar{z}}^T \bar{V}_x(\mathbf{z}^*) \mathbf{\bar{z}},
    \label{eq_lingp_var}
\end{equation}
This LinGP approximation with mean (\ref{eq_lingp_mean}) and variance (\ref{eq_lingp_var}) approximates the mean (\ref{eq_mean}) of the original GP by a linear function of $\mathbf{\bar{z}}$ and the variance (\ref{eq_var}) as a quadratic of $\mathbf{\bar{z}}$. 

It is important to note that LinGP is a GP model of $\tilde{d}_x(\Delta_{\mathbf{z}})$ in (\ref{eq_lin_d_2}), which approximates the original GP around $\mathbf{z}^*$. We will exploit the fact the LinGP has posterior mean and variance that are linear and convex quadratic functions in $\mathbf{\bar{z}}$, respectively. 

We perform the same approximation for the disturbances in the $y$ and $z$ directions, i.e., $d_y(\mathbf{z})$ and $d_z(\mathbf{z})$, independently learnt as GPs. Consequently, we obtain the approximations $\tilde{d}_y(\Delta_{\mathbf{z}})|_{\mathcal{D}} \sim \mathcal{N}(\tilde{\mu}_y(\Delta_{\mathbf{z}}), \tilde{\sigma}^2_y(\Delta_{\mathbf{z}}))$  for $d_y(\mathbf{z})$ and $\tilde{d}_z(\Delta_{\mathbf{z}})|_{\mathcal{D}} \sim \mathcal{N}(\tilde{\mu}_z(\Delta_{\mathbf{z}}), \tilde{\sigma}^2_z(\Delta_{\mathbf{z}}))$ for $d_z(\mathbf{z})$ where 
$\tilde{\mu}_y(\Delta_{\mathbf{z}}) = \bar{\mu}_y(\mathbf{z}^*)^T \mathbf{\bar{z}}$, $\tilde{\sigma}^2_y(\Delta_{\mathbf{z}}) = \mathbf{\bar{z}}^T \bar{V}_y(\mathbf{z}^*) \mathbf{\bar{z}} $, $\tilde{\mu}_z(\Delta_{\mathbf{z}}) = \bar{\mu}_z(\mathbf{z}^*)^T \mathbf{\bar{z}}$, and $\tilde{\sigma}^2_z(\Delta_{\mathbf{z}}) = \mathbf{\bar{z}}^T \bar{V}_z(\mathbf{z}^*) \mathbf{\bar{z}}$. We approximate the disturbance in (\ref{eq_thrust}) with $\mathbf{d}(\mathbf{z}^* + \Delta_{\mathbf{z}}) \approx \mathbf{\tilde{d}}(\Delta_{\mathbf{z}})$ where $\mathbf{\tilde{d}}(\Delta_{\mathbf{z}}) = [\tilde{d}_x(\Delta_{\mathbf{z}}), \tilde{d}_y(\Delta_{\mathbf{z}}), \tilde{d}_z(\Delta_{\mathbf{z}})]^T$. Using the LinGP model, see (\ref{eq_lingp_mean}) - (\ref{eq_lingp_var}), $\mathbf{\tilde{d}}(\Delta_{\mathbf{z}}) |_{\mathcal{D}} \sim \mathcal{N}(\pmb{\tilde{\mu}}(\Delta_{\mathbf{z}}), \pmb{\tilde{\Sigma}}(\Delta_{\mathbf{z}}))$ where:
\begin{equation}
    \pmb{\tilde{\mu}}(\Delta_{\mathbf{z}}) = [\bar{\mu}_x(\mathbf{z}^*), \bar{\mu}_y(\mathbf{z}^*), \bar{\mu}_z(\mathbf{z}^*)]^T \mathbf{\bar{z}},
    \label{eq_mean_d}
\end{equation}
by plugging in (\ref{eq_lingp_mean}) for the mean in the $x$ direction and similar expressions obtained in the $y$ and $z$ directions. The covariance matrix is found from (\ref{eq_lingp_var}) for the $x$ direction and similar expressions obtained in the $y$ and $z$ directions as:
\begin{equation}
    \pmb{\tilde{\Sigma}}(\Delta_{\mathbf{z}}) = \text{diag}(\tilde{\sigma}^2_x(\Delta_{\mathbf{z}}), \tilde{\sigma}^2_y(\Delta_{\mathbf{z}}), \tilde{\sigma}^2_z(\Delta_{\mathbf{z}})),
    \label{eq_var_d}
\end{equation}
where $\text{diag}(\cdot)$ is used to denote a 3$\times$3 diagonal matrix with the elements on its main diagonal. 

\subsection{Second-Order Cone Constraints}
We will reformulate both chance constraints (\ref{eq_ball}) and (\ref{eq_cone}) as second-order cone (SOC) constraints by first reformulating them in terms of the mean disturbance (\ref{eq_mean_d}) and its covariance (\ref{eq_var_d}). To do this, we use the definition of probabilistic reachable sets, an extension of the concept of reachable sets to stochastic systems, see \cite{c8}. 

\begin{definition}
A set $\mathcal{R}$ is said to be a probabilistic i-step reachable set (i-step PRS) of probability level $p_r$ if:
$$ \text{Pr}(\mathbf{e}_i \in \mathcal{R} | \mathbf{e}_0 = 0) \geq p_r. $$
\end{definition}
We define the error at time step $i$ as $\mathbf{e}_i := \mathbf{T}_i - \pmb{\tilde{\mu}}_{\mathbf{T},i}$ where $\mathbf{T}_i$ is the thrust vector in (\ref{eq_thrust}) at time step $i$ and 
\begin{equation}
    \pmb{\tilde{\mu}}_{\mathbf{T},i} = m \mathbf{a}_i + m g \mathbf{z}_W - \pmb{\tilde{\mu}}(\Delta_{\mathbf{z},i}),
    \label{eq_mean_thrust}
\end{equation}
is the posterior mean of the thrust vector at time step $i$. 

Given an $i$-step PRS $\mathcal{R}$ of probability level $p_r$ for the thrust error $\mathbf{e}_i$, if we define tightened constraints, see \cite{c8}, on the mean  $\pmb{\tilde{\mu}}_{\mathbf{T},i}$ as:
\begin{equation}
    \pmb{\tilde{\mu}}_{\mathbf{T},i} \in S \ominus \mathcal{R},
    \label{eq_tighten}
\end{equation}
where $\ominus$ denotes the Pontryagin set difference, then it implies satisfaction of the constraint $\text{Pr}(\mathbf{T}_i \in S) \geq p_r$ where $S$ is a convex set. 
Given that $\mathbf{e}_i \sim \mathcal{N}(0, \pmb{\tilde{\Sigma}_{T}}(\Delta_{\mathbf{z},i}))$ where $\pmb{\tilde{\Sigma}_{T}}(\Delta_{\mathbf{z},i}) = \pmb{\tilde{\Sigma}}(\Delta_{\mathbf{z},i})$ is the covariance matrix (\ref{eq_var_d}) at time step $i$ (a zero-mean normal distribution), the PRS $\mathcal{R}$ can be completely characterized by  $\pmb{\tilde{\Sigma}}(\Delta_{\mathbf{z},i})$, i.e., as ellipsoidal confidence regions as:
\begin{equation}
    \mathcal{R} = \{ \mathbf{e}_i | \quad \mathbf{e}^T_i \pmb{\tilde{\Sigma}}^{-1}(\Delta_{\mathbf{z},i}) \mathbf{e}_i \leq \mathcal{X}^2_3(p_r) \},
    \label{eq_prs}
\end{equation}
where  $\mathcal{X}^2_3(p_r)$ is the quantile function of the chi-squared distribution with three degrees of freedom. 
Despite this, the online computation of the tightening in (\ref{eq_tighten}) is often computationally prohibitive for general convex constraint sets $S$. However, we will exploit the fact that our convex sets $S$ comprise of a ball $S_{\text{ball}}$ in (\ref{eq_ball}) and cone $S_{\text{cone}}$ in (\ref{eq_cone}) to compute (\ref{eq_tighten}) in an efficient way that can be implemented online in (\ref{eq_fmpc}) to solve the optimal control problem as a second-order cone program (SOCP). 

\subsubsection{Ball Constraint Tightening} To simplify the constraint tightening $S_{\text{ball}} \ominus \mathcal{R}$ in (\ref{eq_tighten}), we use an outer approximation of the PRS (\ref{eq_prs}) by a ball $ \mathcal{R} \subseteq \mathcal{R}_b$  as:
\begin{equation}
    \mathcal{R}_{b} = \left\{ \mathbf{e}_i | \quad ||\mathbf{e}_i||_2 \leq \sqrt{\lambda_{\text{max}}(\pmb{\tilde{\Sigma}}(\Delta_{\mathbf{z},i})) \mathcal{X}^2_3(p_r)} \right\},
    \label{eq_prs_ball}
\end{equation}
where $\lambda_{\text{max}}(\cdot)$ is  the  maximum  eigenvalue. 

We will now compute the tightened constraint (\ref{eq_tighten}) as $S_{\text{ball}} \ominus \mathcal{R}_{b}$. To do this, we use the triangle inequality and (\ref{eq_prs_ball}) where the thrust vector at time step $i$ is upper bounded as $|| \mathbf{T}_i ||_2 = || \pmb{\tilde{\mu}}_{\mathbf{T},i} + \mathbf{e}_i ||_2 \leq || \pmb{\tilde{\mu}}_{\mathbf{T},i}||_2 + || \mathbf{e}_i ||_2 \leq || \pmb{\tilde{\mu}}_{\mathbf{T},i}||_2 + \sqrt{\lambda_{\text{max}}(\pmb{\tilde{\Sigma}}(\Delta_{\mathbf{z},i})) \mathcal{X}^2_3(p_r)}$. Consequently, (\ref{eq_ball}) holds if:
\begin{equation}
 ||\pmb{\tilde{\mu}}_{\mathbf{T},i}||_2 \leq T_{max} - \sqrt{\lambda_{\text{max}}(\pmb{\tilde{\Sigma}}(\Delta_{\mathbf{z},i}))\mathcal{X}^2_3(p_b)}, 
 \label{eq_ball_tight}
\end{equation}
where $p_b = p_r$.

Using (\ref{eq_mean_d}) and (\ref{eq_mean_thrust}), we can write the posterior mean thrust  $\pmb{\tilde{\mu}}_{\mathbf{T},i}$ at time step $i$ as a linear function of the augmented vector $\mathbf{\bar{z}}_i = [1 \quad \Delta_{\mathbf{z}_i} ]^T$ at time step $i$ :
\begin{equation}
    \pmb{\tilde{\mu}}_{\mathbf{T},i} = \mathbf{h}_i^T \mathbf{\bar{z}}_i.
    \label{eq_linear}
\end{equation}
The covariance $\pmb{\tilde{\Sigma}}(\Delta_{\mathbf{z},i})$ at time step $i$ is a diagonal matrix in (\ref{eq_var_d}) and, therefore, the maximum eigenvalue is either $\tilde{\sigma}^2_x(\Delta_{\mathbf{z}_i})$, $\tilde{\sigma}^2_y(\Delta_{\mathbf{z}_i})$ or $\tilde{\sigma}^2_z(\Delta_{\mathbf{z}_i})$ at time step $i$. We enforce (\ref{eq_ball_tight}) through three constraints:
\vspace{-2mm}
\begin{equation}
\begin{aligned}
     ||\pmb{\tilde{\mu}}_{\mathbf{T},i}||_2 & \leq T_{max} - \sqrt{ \tilde{\sigma}^2_x(\Delta_{\mathbf{z}_i}) \mathcal{X}^2_3(p_b)}, \\
     ||\pmb{\tilde{\mu}}_{\mathbf{T},i}||_2 & \leq T_{max} - \sqrt{ \tilde{\sigma}^2_y(\Delta_{\mathbf{z}_i}) \mathcal{X}^2_3(p_b)}, \\
     ||\pmb{\tilde{\mu}}_{\mathbf{T},i}||_2 & \leq T_{max} - \sqrt{ \tilde{\sigma}^2_z(\Delta_{\mathbf{z}_i}) \mathcal{X}^2_3(p_b)}.
\end{aligned}
\label{eq_3_constraints}
\end{equation}
We now exploit the quadratic variance (\ref{eq_lingp_var}) in augmented vector $\mathbf{\bar{z}}_i = [1 \quad \Delta_{\mathbf{z}_i} ]^T$ at time step $i$. Then, $\sqrt{\tilde{\sigma}^2_x(\Delta_{\mathbf{z}_i})} = \sqrt{\mathbf{\bar{z}}_i^T \bar{V}_x(\mathbf{z}_i^*) \mathbf{\bar{z}}_i} = || \bar{L}^T_x(\mathbf{z}_i^*) \mathbf{\bar{z}}_i ||_2$ where  $\bar{L}_x(\mathbf{z}_i^*)$ is the Cholesky decomposition of covariance matrix $\bar{V}_x(\mathbf{z}_i^*)$ in (\ref{eq_cov_matrix}), i.e., $\bar{V}_x(\mathbf{z}_i^*) = \bar{L}_x(\mathbf{z}_i^*) \bar{L}^T_x(\mathbf{z}_i^*)$ where $\bar{L}_x(\mathbf{z}_i^*)$ is a real lower triangular matrix with positive diagonal entries. Note, we can do this because the covariance matrix $\bar{V}_x(\mathbf{z}_i^*)$ will always be symmetric and positive definite. Similarly, for the $y$ and $z$ directions, $\sqrt{\tilde{\sigma}^2_y(\Delta_{\mathbf{z}_i})} = || \bar{L}^T_y(\mathbf{z}_i^*) \mathbf{\bar{z}}_i ||_2$ and $\sqrt{\tilde{\sigma}^2_z(\Delta_{\mathbf{z}_i})} = ||\bar{L}^T_z(\mathbf{z}_i^*) \mathbf{\bar{z}}_i||_2$ where $\bar{L}_y(\mathbf{z}_i^*)$ and $\bar{L}_z(\mathbf{z}_i^*)$ are the Cholesky decomposition of covariance matrices $\bar{V}_y(\mathbf{z}_i^*)$ and $\bar{V}_z(\mathbf{z}_i^*)$ respectively. Plugging in (\ref{eq_linear}) and using the Cholesky decomposition in (\ref{eq_3_constraints}), we obtain: 
%\vspace{-2mm}
\begin{equation}
\begin{aligned}
     ||\mathbf{h}_i^T \mathbf{\bar{z}}_i||_2 & \leq T_{max} - \sqrt{\mathcal{X}^2_3(p_b)} || \bar{L}^T_x(\mathbf{z}_i^*) \mathbf{\bar{z}}_i ||_2, \\
     ||\mathbf{h}_i^T \mathbf{\bar{z}}_i||_2 & \leq T_{max} - \sqrt{\mathcal{X}^2_3(p_b)} || \bar{L}^T_y(\mathbf{z}_i^*) \mathbf{\bar{z}}_i ||_2, \\
     ||\mathbf{h}_i^T \mathbf{\bar{z}}_i||_2 & \leq T_{max} - \sqrt{\mathcal{X}^2_3(p_b)} || \bar{L}^T_z(\mathbf{z}_i^*) \mathbf{\bar{z}}_i||_2.
\end{aligned}
\label{eq_3_soc_constraints}
\end{equation}
We introduce three intermediate variables $\alpha_{1,i}, \alpha_{2,i}, \alpha_{3,i}$ and rewrite (\ref{eq_3_soc_constraints}) as six SOC constraints on $\mathbf{\bar{z}}_i$ and variables $\alpha_{1,i}, \alpha_{2,i}, \alpha_{3,i}$ at each time step $i$ as:
\vspace{-2mm}
\begin{equation}
\begin{aligned}
   & ||\mathbf{h}_i^T \mathbf{\bar{z}}_i||_2 \leq T_{max} - \sqrt{\mathcal{X}^2_3(p_b)} \alpha_{1,i}, \\
    & ||\mathbf{h}_i^T \mathbf{\bar{z}}_i||_2 \leq T_{max} - \sqrt{\mathcal{X}^2_3(p_b)} \alpha_{2,i}, \\
    & ||\mathbf{h}_i^T \mathbf{\bar{z}}_i||_2 \leq T_{max} - \sqrt{\mathcal{X}^2_3(p_b)} \alpha_{3,i},\\
   & || \bar{L}^T_x(\mathbf{z}_i^*) \mathbf{\bar{z}}_i ||_2  \leq \alpha_{1,i}, \\
   & || \bar{L}^T_y(\mathbf{z}_i^*) \mathbf{\bar{z}}_i ||_2 \leq \alpha_{2,i}, \\
   & || \bar{L}^T_z(\mathbf{z}_i^*) \mathbf{\bar{z}}_i ||_2 \leq \alpha_{3,i}. \\
\end{aligned}
\label{eq_6_soc_constraints}
\end{equation}

We have now described the ball constraint (\ref{eq_ball}) as six SOC constraints in (\ref{eq_6_soc_constraints}). We will use a similar approach to obtain SOC constraints for the cone constraint (\ref{eq_cone}). 

\subsubsection{Cone Constraint Tightening} We rewrite the cone constraint (\ref{eq_cone}) by introducing an intermediate variable $r$ such that:
\begin{equation}
    \text{Pr}\{ || [T_x \quad T_y]^T ||_2 \leq r \} \geq p_{c,1},
    \label{eq_cone_1}
\end{equation}
\begin{equation}
    \text{Pr}\{ r \leq \tan{\theta_{max}} T_z \} \geq p_{c,2},
    \label{eq_cone_2}
\end{equation}
where $p_{c,1} p_{c,2} \geq p_c$. We will treat the tightening of the ball $S_{\text{ball,c}}$ described by constraint ($\ref{eq_cone_1}$) and half-space $S_{\text{half,c}}$ described by constraint (\ref{eq_cone_2}) independently. 

To tighten constraint ($\ref{eq_cone_1}$) at time step $i$, i.e., $S_{\text{ball,c}} \ominus \mathcal{R}$, we follow a similar procedure to tightening (\ref{eq_ball}). We obtain two constraints:
\vspace{-2mm}
\begin{equation}
\begin{aligned}
     ||\mathbf{h}_c \mathbf{h}_i^T \mathbf{\bar{z}}_i||_2 & \leq r_i - \sqrt{\mathcal{X}^2_3(p_{c,1})} || \bar{L}^T_x(\mathbf{z}_i^*) \mathbf{\bar{z}}_i ||_2, \\
     ||\mathbf{h}_c \mathbf{h}_i^T \mathbf{\bar{z}}_i||_2 & \leq r_i - \sqrt{\mathcal{X}^2_3(p_{c,1})} || \bar{L}^T_y(\mathbf{z}_i^*) \mathbf{\bar{z}}_i ||_2,
\end{aligned}
\label{eq_2_soc_constraints}
\end{equation}
where $\mathbf{h}_c = [1 \quad 1 \quad 0]$ selects out $x$ and $y$ components of the posterior mean thrust $\pmb{\tilde{\mu}}_{\mathbf{T},i}$ in  (\ref{eq_linear}). We introduce two intermediate variables $\beta_{1,i}$, $\beta_{2,i}$ and rewrite (\ref{eq_2_soc_constraints}) as four SOC constraints on $\mathbf{\bar{z}}_i$ and variables $\beta_{1,i}, \beta_{2,i}, r_{i}$ at each time step $i$ as:
\vspace{-2mm}
\begin{equation}
\begin{aligned}
    & ||\mathbf{h}_c \mathbf{h}_i^T \mathbf{\bar{z}}_i||_2 \leq r_i - \sqrt{\mathcal{X}^2_3(p_{c,1})} \beta_{1,i},\\
    & ||\mathbf{h}_c \mathbf{h}_i^T \mathbf{\bar{z}}_i||_2 \leq r_i - \sqrt{\mathcal{X}^2_3(p_{c,1})} \beta_{2,i}, \\
    & || \bar{L}^T_x(\mathbf{z}_i^*) \mathbf{\bar{z}}_i ||_2 \leq \beta_{1,i}, \\
    & || \bar{L}^T_y(\mathbf{z}_i^*) \mathbf{\bar{z}}_i ||_2 \leq \beta_{2,i}.
\end{aligned}
\label{eq_4_soc_constraints}
\end{equation}
At each time step, we need to satisfy both the SOC constraints (\ref{eq_6_soc_constraints}) and (\ref{eq_4_soc_constraints}). To tighten constraint ($\ref{eq_cone_2}$) at time step $i$, i.e., $S_{\text{half,c}} \ominus \mathcal{R}$, we first rewrite (\ref{eq_cone_2}) as a half-space constraint on $\pmb{\zeta}_i = [r_i \quad T_{z,i}]^T$  as:
\vspace{-2mm}
\begin{equation}
    \text{Pr} \{ \mathbf{h}^T_h \pmb{\zeta}_i \leq 0 \} \geq p_{c,2}
    \label{eq_half}
\end{equation}
where $\mathbf{h}_h = [1 \quad -\tan(\theta_{max})]^T$. We define
\begin{equation}
    \mathbf{e}_{\pmb{\zeta},i} := \pmb{\zeta}_i - [r_i \quad \pmb{\lambda} \pmb{\tilde{\mu}}_{\mathbf{T},i} ]^T
    \label{eq_error}
\end{equation}
where $\pmb{\lambda} = [0, 0, 1]$ extracts the posterior mean thrust in the $z$ direction. Given that $\mathbf{e}_{\pmb{\zeta},i} \sim \mathcal{N}(0, \pmb{\Sigma}_{\pmb{\zeta}}(\Delta_{\mathbf{z}_i}))$ where $\pmb{\Sigma}_{\pmb{\zeta}}(\Delta_{\mathbf{z}_i}) = \text{diag}(0, \tilde{\sigma}^2_z(\Delta_{\mathbf{z}_i})))$, it follows that $\mathbf{h}^T_h \mathbf{e}_{\pmb{\zeta},i} \sim \mathcal{N}(0, \mathbf{h}^T_h\pmb{\Sigma}_{\pmb{\zeta}}(\Delta_{\mathbf{z}_i})\mathbf{h}_h)$. The PRS $\mathcal{R}_h$ with probability level $p_{c,2}$ of $\mathbf{h}^T_h \mathbf{e}_{\pmb{\zeta},i}$ is:
\begin{equation}
    \mathcal{R}_h = \left\{ \mathbf{e}_{\pmb{\zeta},i} | \mathbf{h}^T_h \mathbf{e}_{\pmb{\zeta},i} \leq \phi^{-1}(p_{c,2}) \sqrt{\mathbf{h}^T_h\pmb{\Sigma}_{\pmb{\zeta}}(\Delta_{\mathbf{z}_i})\mathbf{h}_h} \right\},
    \label{eq_r_h}
\end{equation}
where $\phi^{-1}(p_{c,2})$ is the quantile function of a standard Gaussian random variable at the needed probability of constraint satisfaction $p_{c,2}$. Using the error (\ref{eq_error}) and (\ref{eq_r_h}), we can rewrite (\ref{eq_half}) as:
\begin{equation}
    \mathbf{h}^T_h [r_i \quad \pmb{\lambda} \pmb{\tilde{\mu}}_{\mathbf{T},i} ]^T \leq - \phi^{-1}(p_{c,2}) \sqrt{\mathbf{h}^T_h\pmb{\Sigma}_{\pmb{\zeta}}(\Delta_{\mathbf{z}_i})\mathbf{h}_h}. 
    \label{eq_half2}
\end{equation}
Using the quadratic variance (\ref{eq_lingp_var}) and plugging in $\mathbf{h}_h$ and $\pmb{\Sigma}_{\pmb{\zeta}}(\Delta_{\mathbf{z}_i})$, it simplifies $\sqrt{\mathbf{h}^T_h\pmb{\Sigma}_{\pmb{\zeta}}(\Delta_{\mathbf{z}_i})\mathbf{h}_h} = \tan(\theta_{max}) \sqrt{\tilde{\sigma}^2_z(\Delta_{\mathbf{z}_i})} = \tan(\theta_{max}) \sqrt{\mathbf{\bar{z}}_i^T \bar{V}_z(\mathbf{z}_i^*) \mathbf{\bar{z}}_i} = \tan(\theta_{max}) ||\bar{L}^T_z(\mathbf{z}_i^*)\mathbf{\bar{z}}_i||_2$ using the Cholesky decomposition of covariance matrix $\bar{V}_z(\mathbf{z}_i^*) = \bar{L}_z(\mathbf{z}_i^*) \bar{L}^T_z(\mathbf{z}_i^*)$. Using this and plugging in (\ref{eq_linear}) into (\ref{eq_half2}) $ \mathbf{h}^T_h [r_i \quad \pmb{\lambda} \mathbf{h}_i^T \mathbf{\bar{z}}_i ]^T \leq - \phi^{-1}(p_{c,2})\tan(\theta_{\text{max}})|| \bar{L}^T_z(\mathbf{z}_i^*) \mathbf{\bar{z}}_i||_2 $ or equivalently:
\begin{equation}
   || \bar{L}^T_z(\mathbf{z}_i^*) \mathbf{\bar{z}}_i||_2 \leq \frac{-1}{\tan(\theta_{max})\phi^{-1}(p_{c,2})} \mathbf{h}^T_h [r_i \quad \pmb{\lambda} \mathbf{h}_i^T \mathbf{\bar{z}}_i ]^T 
   \label{eq_1_soc_constraint}
\end{equation}
which is a SOC constraint on $\mathbf{\bar{z}}_i$ and variable $r_{i}$ at time step $i$. At each time step $i$, our feasibility constraints consist of eleven SOC constraints represented by (\ref{eq_6_soc_constraints}), (\ref{eq_4_soc_constraints}) and (\ref{eq_1_soc_constraint}).

\subsection{Solving MPC as Second-Order Cone Program}

Our proposed MPC (Learning SOCP) takes the current flat state $\mathbf{z}_i = \mathbf{z}_{\text{init}}$ at each time step $i$ 
and optimizes for a sequence of augmented states $\mathbf{\bar{z}}_{0:N}$, see (\ref{eq_aug}) for definition, a sequence of control commands $\mathbf{v}_{0:N-1}$, a sequence of intermediate (or dummy) variables $\alpha_{1,0:N}, \alpha_{2,0:N}, \alpha_{3,0:N}$ (used at each time step along the prediction horizon in (\ref{eq_6_soc_constraints})), a sequence of intermediate variables $\beta_{1,0:N}, \beta_{2,0:N}$ (used at each time step along the prediction horizon in (\ref{eq_4_soc_constraints})) and a sequence of intermediate variable $r_{0:N}$ (used in (\ref{eq_4_soc_constraints}) and (\ref{eq_1_soc_constraint})). We use the same quadratic cost as FMPC (\ref{eq_cost}). Note that the cost (\ref{eq_cost}) is still quadratic in terms of augmented state $\mathbf{\bar{z}}_{0:N}$ in (\ref{eq_aug}). We convert the quadratic cost into two second-order cones. To do this and write our optimal control problem (OCP) in (\ref{eq_fmpc}) subject to constraints (\ref{eq_ball})-(\ref{eq_cone}) as a SOCP, we introduce additional optimization variables $\gamma_{1,0:N}$ and $\gamma_{2,0:{N-1}}$ such that:
\begin{equation}
    (\mathbf{C} \mathbf{{z}}_k - \mathbf{y}^{ref}_k)^T\mathbf{Q}(\mathbf{C} \mathbf{{z}}_k - \mathbf{y}^{ref}_k)  \leq \gamma_{1,k} \quad \forall k = 1, ..., N 
    \label{eq_soc_cost1}
\end{equation}
\begin{equation}
\mathbf{v}_k^T \mathbf{R} \mathbf{v}_k \leq \gamma_{2,k} \quad \forall k = 0, ..., N-1 
\label{eq_soc_cost2}
\end{equation}
Use the fact that $\mathbf{Q} \succ 0$, we can rewrite (\ref{eq_soc_cost1}) as a SOC: 
\begin{equation}
         \newcommand{\norm}[1]{\left\lVert#1\right\rVert}\norm{\begin{bmatrix}\mathbf{Q}^{\frac{1}{2}}(\mathbf{C} \mathbf{z}_k - \mathbf{y}^{ref}_k) \\ 1 -  \gamma_{1,k}\end{bmatrix}}_2 \leq \gamma_{1,k} + 1 \quad \forall k = 1, ..., N 
         \label{eq_soc_cost11}
\end{equation}
Similarly, using $\mathbf{R} \succ 0$, we can rewrite (\ref{eq_soc_cost2}) as a SOC: 
\begin{equation}
         \newcommand{\norm}[1]{\left\lVert#1\right\rVert}\norm{\begin{bmatrix}\mathbf{R}^{\frac{1}{2}}\mathbf{v}_k \\ 1 -  \gamma_{2,k}\end{bmatrix}}_2 \leq \gamma_{2,k} + 1 \quad \forall k = 0, ..., N-1 
         \label{eq_soc_cost22}
\end{equation}
Our proposed MPC solves the optimal control problem (OCP) using the discretized linear dynamics (\ref{eq_lin_flat}) as a SOCP:
\vspace{-2mm}
\begin{equation}
\begin{aligned}
	\min & \sum_{k=1}^{N} \gamma_{1,k} + \gamma_{2,{k-1}} \\
\textrm{s.t.} \quad & \mathbf{z}_{k+1} = \mathbf{A}_d \mathbf{z}_k + \mathbf{B}_d \mathbf{v}_k \quad \forall k = 0,..., N-1\\
& \Delta_{\mathbf{z}, k} = \mathbf{z}_k - \mathbf{z}^{*}_k \quad \forall k = 1,..., N \\
& \mathbf{\bar{z}}_{k} = [1 \quad \Delta_{\mathbf{z}, k}]^T \quad \forall k = 1,..., N \\
& \text{SOC constraints } (\ref{eq_6_soc_constraints}), (\ref{eq_4_soc_constraints}), (\ref{eq_1_soc_constraint}) \quad \forall k = 1,..., N \\
& \text{SOC constraints } (\ref{eq_soc_cost11}), (\ref{eq_soc_cost22}) \\
  &\mathbf{z}_{0} = \mathbf{z}_{\text{init}}
\end{aligned}
\label{eq_socp}
\end{equation}
where $\mathbf{z}^{*}_{\text{traj}} = [\mathbf{z}^{*}_1, ..., \mathbf{z}^{*}_N]$ is the previous optimal trajectory (solving (\ref{eq_socp}) at time step $i-1$)  that we use in LinGP as shown in Fig. 1. 
By optimizing (\ref{eq_socp}), our proposed MPC is predicting an optimal trajectory at each time step where it is possible to determine a feasible thrust command (magnitude and orientation) at each point along that trajectory to compensate for drag. To execute this trajectory, however, we will need to compensate or adapt to the drag disturbance. 

\begin{figure*}
 \centering
 \subfigure[Absolute Path Error]{
      \centering
      \includegraphics[width=0.45\textwidth, trim={0cm 0cm 1cm 1cm},clip]{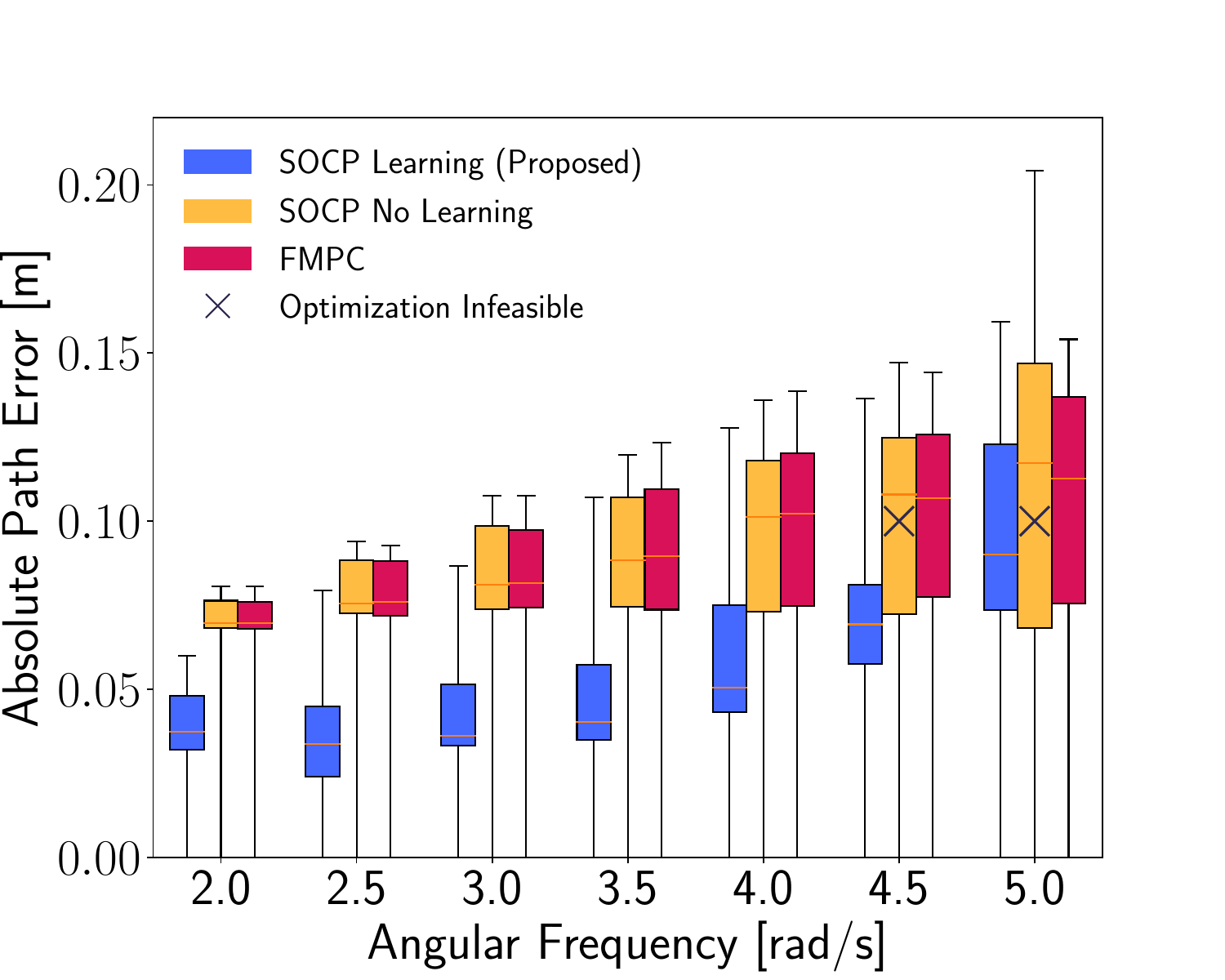}
		} 
 \subfigure[Thrust Angle]{
      \centering
      \includegraphics[width=0.45\textwidth, trim={0cm 0cm 1cm 1cm},clip]{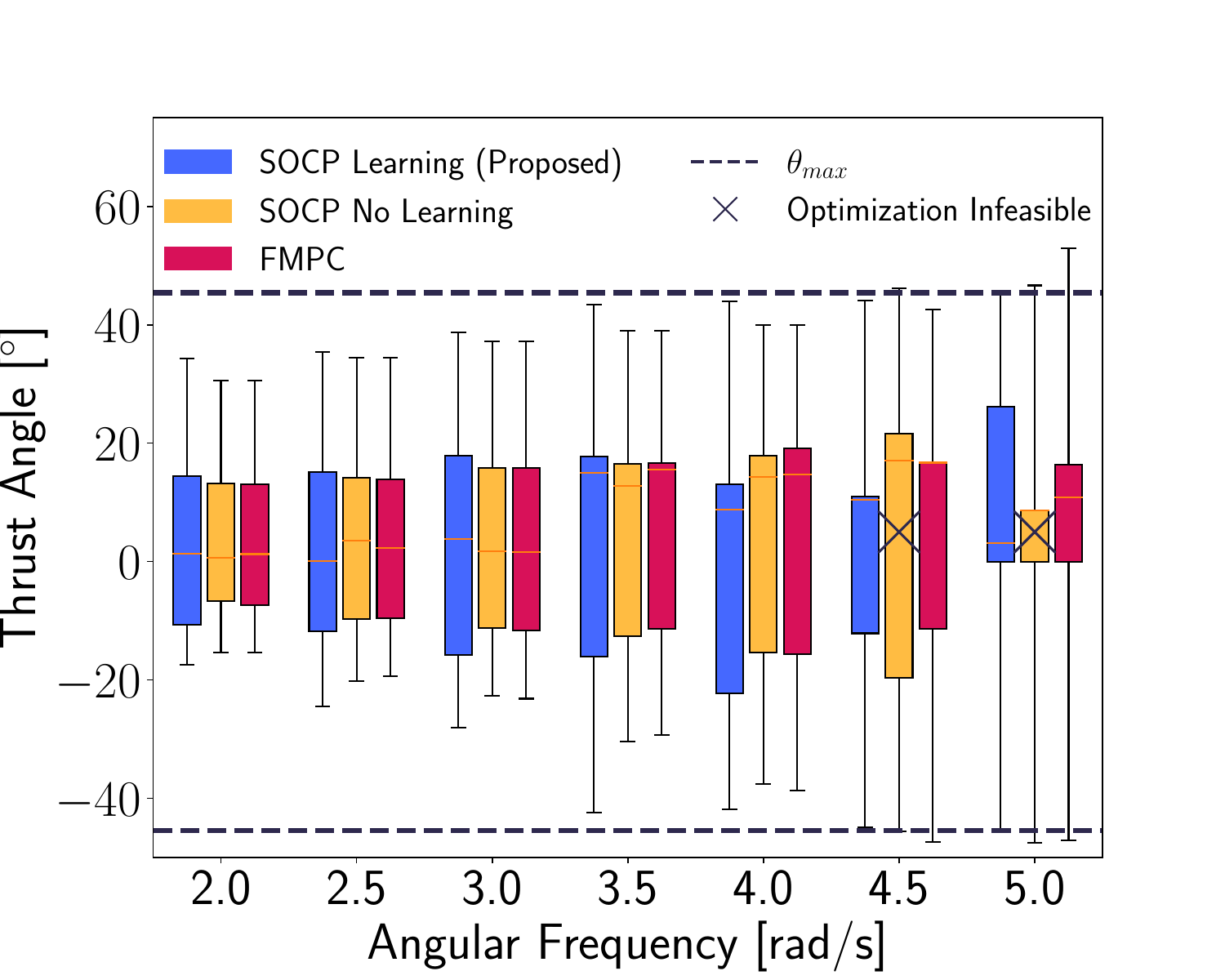}
		} 
 \caption{Comparison of performance and input feasibility for multirotor subject to \textit{linear drag}. We compare the (a) absolute path error and (b) thrust angle for increasingly aggressive sinusoidal trajectories using FMPC (red), SOCP No Learning (yellow) and our proposed SOCP Learning (blue). Our proposed approach outperforms similar methods that neglect the effects of drag on feasibility. The optimization for SOCP No Learning (yellow) goes infeasible for trajectories $\omega = 4.5$ and  $\omega = 5.0$ (marked with a $\times$). } 
 \label{fig:lindrag}
 \vspace{-5mm}
\end{figure*}

\begin{figure*}
 \centering
 \subfigure[Absolute Path Error]{
      \centering
      \includegraphics[width=0.45\textwidth, trim={0cm 0cm 1cm 1cm},clip]{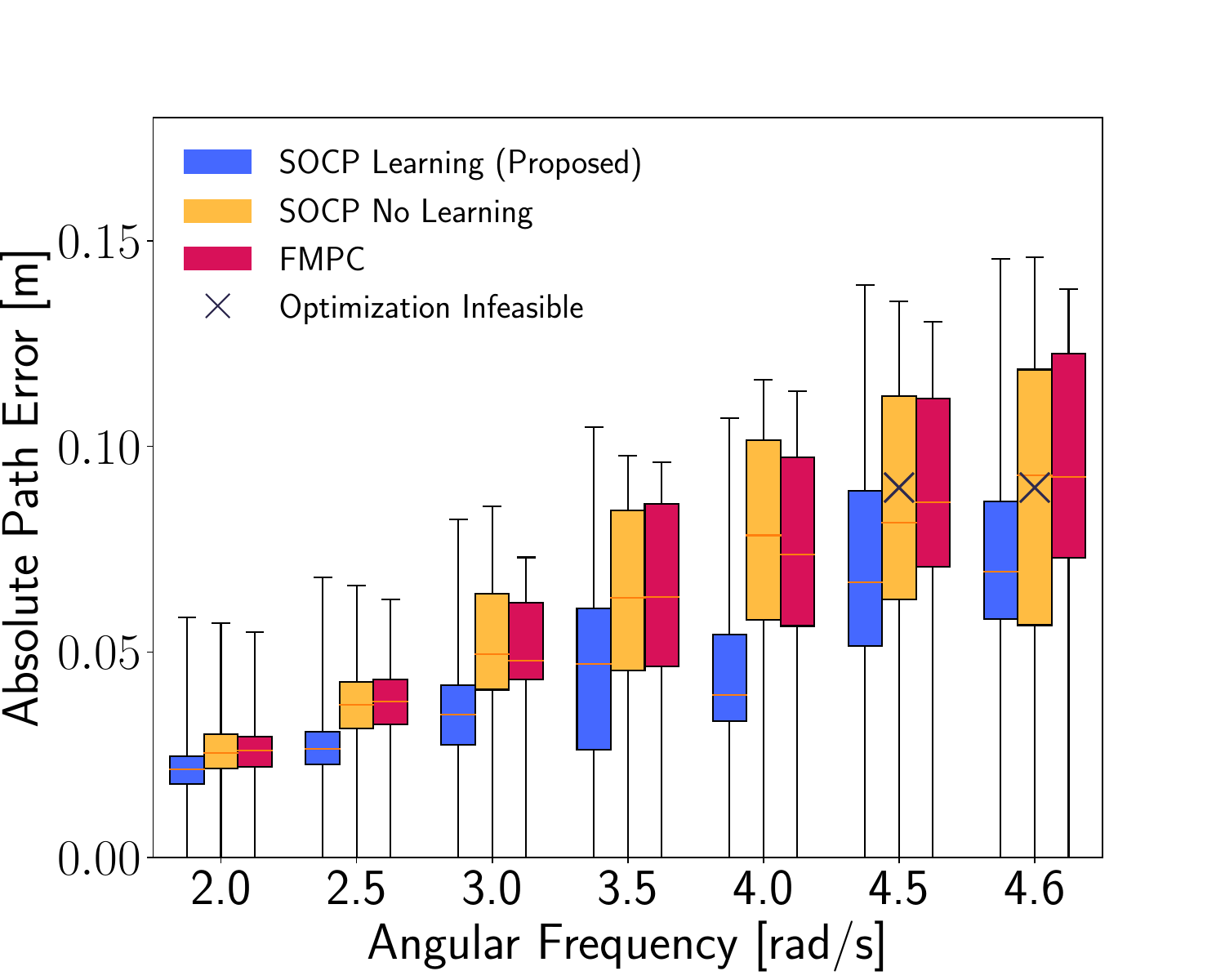}
		} 
 \subfigure[Thrust Angle]{
      \centering
      \includegraphics[width=0.45\textwidth, trim={0cm 0cm 1cm 1cm},clip]{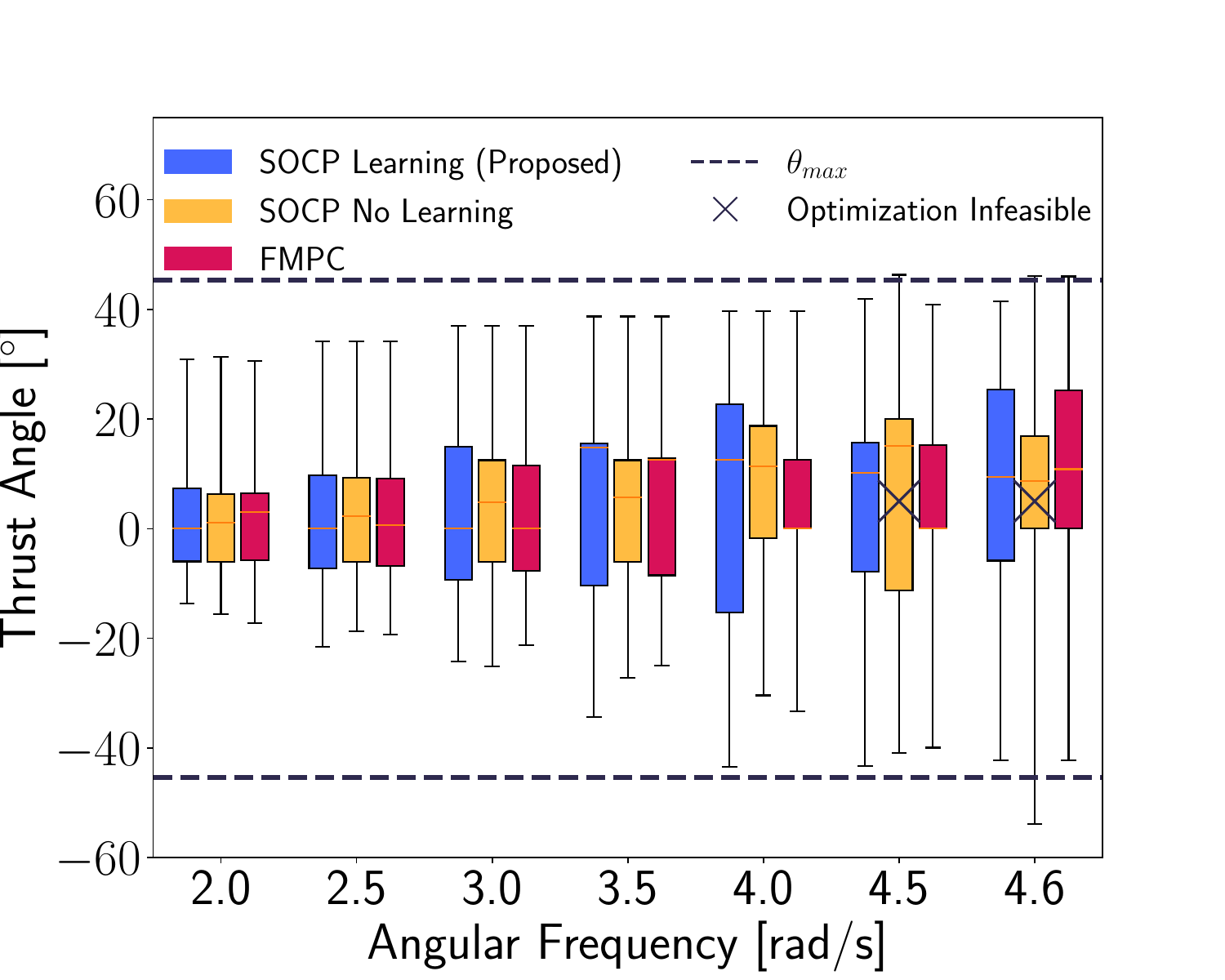}
		} 
 \caption{Comparison of performance and input feasibility for multirotor subject to \textit{quadratic drag}. We compare the (a) absolute path error and (b) thrust angle for increasingly aggressive sinusoidal trajectories using FMPC (red), SOCP No Learning (yellow) and our proposed SOCP Learning (blue). Our proposed approach outperforms similar methods that neglect the effects of drag on feasibility. The optimization for SOCP No Learning (yellow) goes infeasible for trajectories $\omega = 4.5$ and  $\omega = 4.6$ (marked with a $\times$).} 
 \label{fig:squareddrag}
 \vspace{-5mm}
\end{figure*}

The first step of the optimized feasible trajectory $\mathbf{z}_d = \mathbf{z}^*_1$ and $\mathbf{v}_d = \mathbf{v}^*_0$ is set as the desired reference. It is possible to use lower-level adaptive controllers to compensate for the drag, e.g. \cite{c20}. We adopt an alternative simple strategy using our learned disturbance, that is, we send a thrust command $\mathbf{T}_d$ accounting for the expected disturbance (or mean) in (\ref{eq_mean_d}), $
    \mathbf{T}_d = m \mathbf{a}_d + mg \mathbf{z}_W - \pmb{\tilde{\mu}}(\Delta_{\mathbf{z}_d})$, to low-level attitude controllers. The yaw command comes directly from MPC as $\dot{\psi}_d$ where $\mathbf{v}_d = [\mathbf{s}_d, \dot{\psi}_d]$.

\section{SIMULATION RESULTS}
We simulate a 2D multirotor with dynamics similar to (\ref{eq_uav_system}) moving in the $x-z$ plane. We compare three related approaches: Flatness-based MPC (\textit{FMPC}) from \cite{c13} with no input constraints, our proposed second-order cone program with thrust vector constraints neglecting the effects of drag (\textit{SOCP No Learning}) and our proposed approach (\textit{SOCP Learning}). We consider a circular reference path $\mathbf{y}^{ref} = [0.3\sin(\omega t), 0.3\cos(\omega t)]$ with increasing angular frequencies $\omega$. 

\begin{figure*}
 \centering
 \subfigure[Path Visualization]{
      \centering
      \includegraphics[width=0.25\textwidth, trim={0cm 0cm 1cm 1cm},clip]{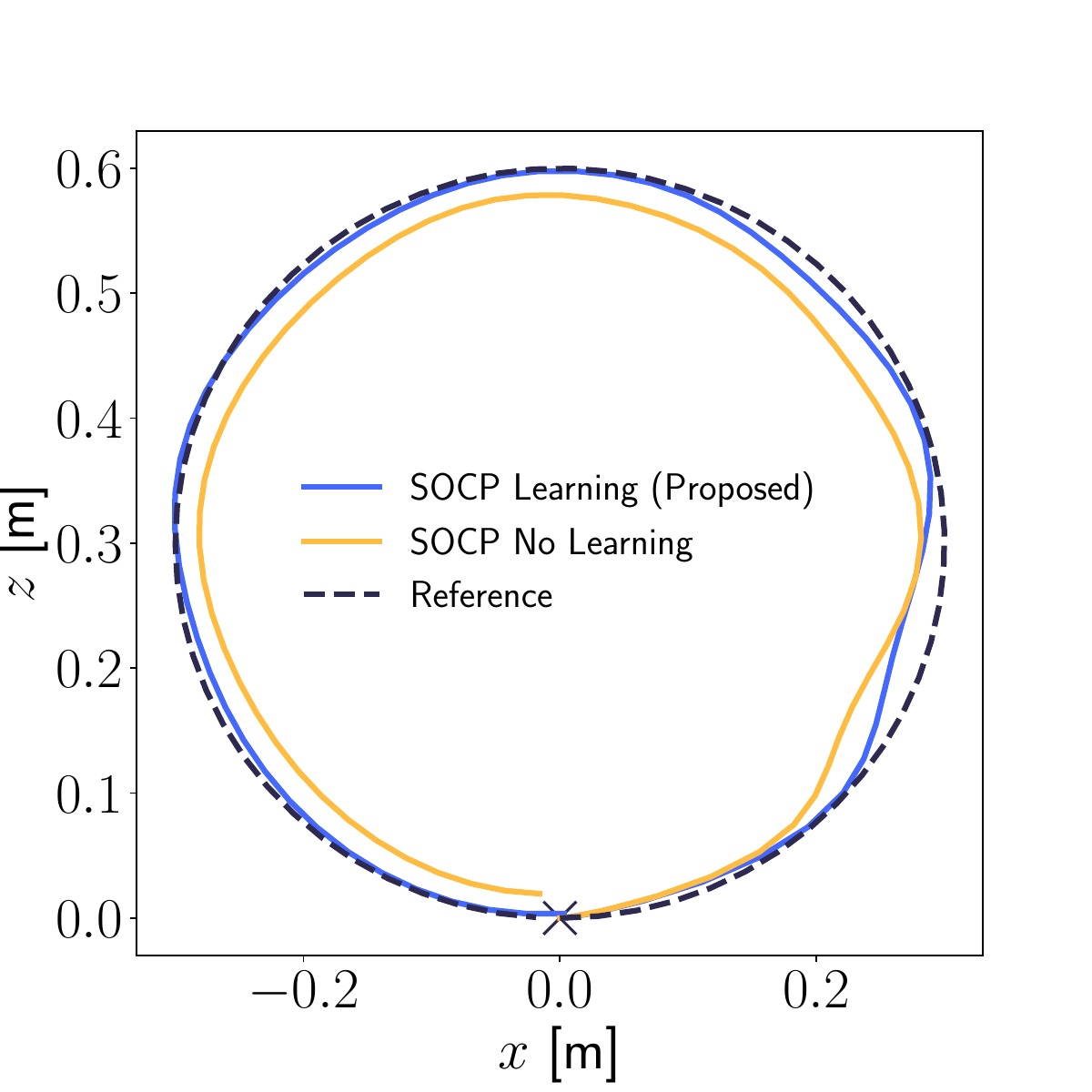}
		} 
 \subfigure[SOCP No Learning]{
      \centering
      \includegraphics[width=0.34\textwidth,  trim={0cm 0cm 1cm 1cm},clip]{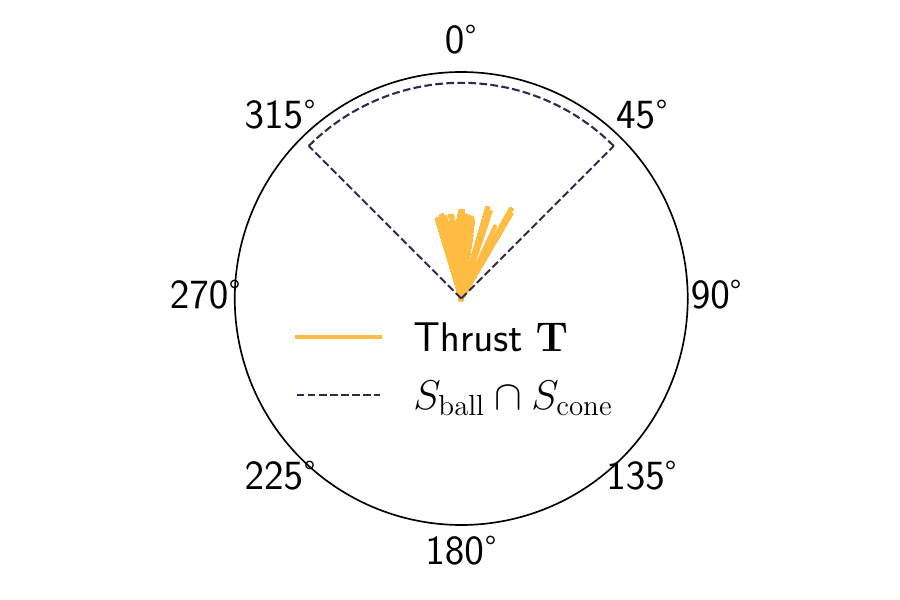}
		} 
   \subfigure[SOCP Learning (Proposed)]{
      \centering
      \includegraphics[width=0.34\textwidth,  trim={0cm 0cm 1cm 1cm},clip]{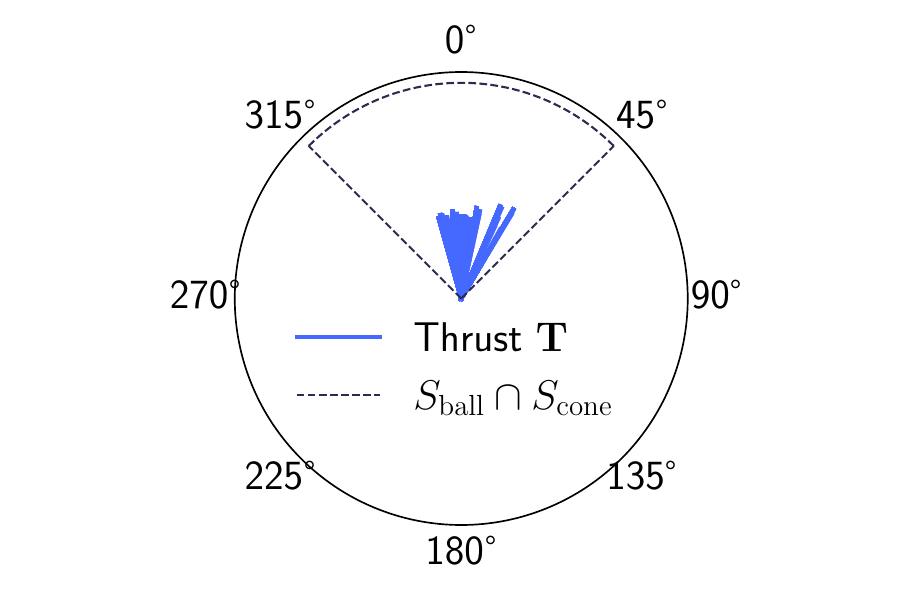}
		} 
 \caption{Visualization of multirotor following sinusoidal trajectory with $\omega = 2$ rad/s  subject to \textit{quadratic drag}: (a) Comparison of path flown under SOCP No Learning (yellow) and our proposed SOCP Learning (blue); Thrust $\mathbf{T}$ and constraints for (b) SOCP No Learning and (c) SOCP Learning (proposed). We observe that our proposed approach sends larger feasible thrust commands to compensate for drag leading to reduced tracking errors. } 
 \vspace{-5mm}
 \label{fig:squareddrag2}
\end{figure*}

\begin{figure*}
 \centering
 \subfigure[Path Visualization]{
      \centering
      \includegraphics[width=0.25\textwidth, trim={0cm 0cm 1cm 1cm},clip]{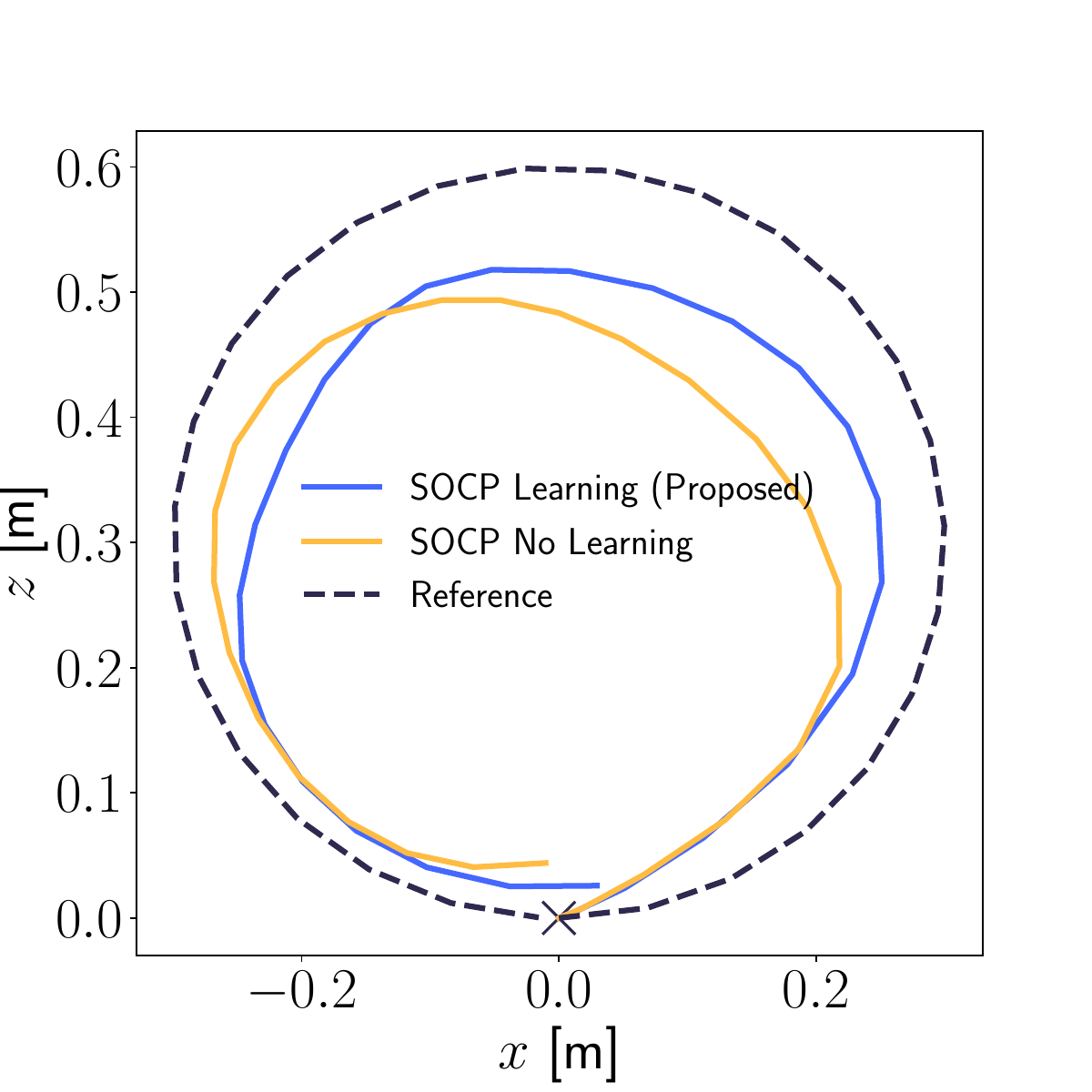}
		} 
 \subfigure[SOCP No Learning]{
      \centering
      \includegraphics[width=0.34\textwidth,  trim={0cm 0cm 1cm 1cm},clip]{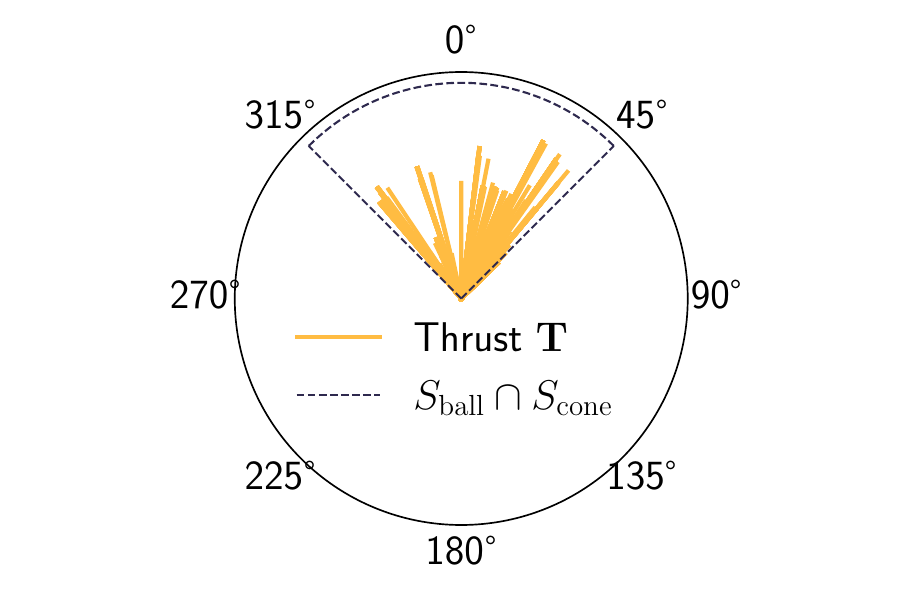}
		} 
   \subfigure[SOCP Learning (Proposed)]{
      \centering
      \includegraphics[width=0.34\textwidth,  trim={0cm 0cm 1cm 1cm},clip]{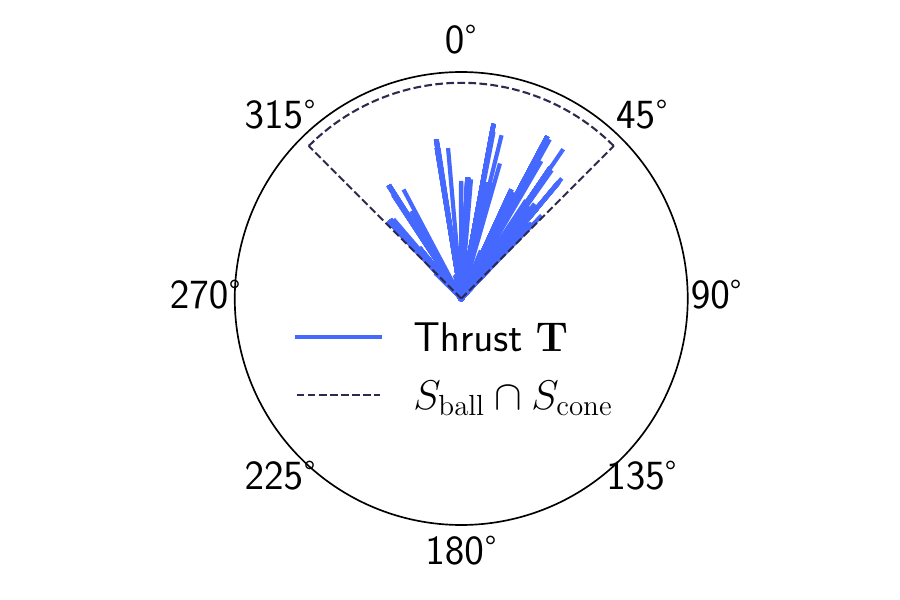}
		} 
  \caption{Visualization of multirotor following sinusoidal trajectory with $\omega = 4.6$ rad/s  subject to \textit{quadratic drag}: (a) Comparison of path flown under SOCP No Learning (yellow) and our proposed SOCP Learning (blue); Thrust $\mathbf{T}$ and constraints for (b) SOCP No Learning and (c) SOCP Learning (proposed). We observe that our proposed approach adapts the trajectory to aerodynamic forces to ensure feasible thrust commands. Neglecting aerodynamic forces leads to constraint violation and infeasibility of the optimal control problem. } 
 \label{fig:squareddrag4.6}
 \vspace{-4mm}
\end{figure*}

We leverage 20 data points in training the GPs of the drag force in (\ref{eq_thrust}) which have been sampled from the trajectory flown using \textit{SOCP No Learning} using Latin hypercube sampling. We use a standard SE kernel and the hyperparameters are optimized to minimize the log-likelihood. The SOCP formulations are solved using MOSEK Fusion API for C++ \cite{c20}. We consider the constraints $\theta_{max} = \frac{\pi}{4}$ rad and $T_{max} = 30$ N. All controllers consider a sample rate of 20 Hz and a look-ahead time of 0.5 s, where the prediction horizon is $N = 10$. The cost matrices $\mathbf{Q} = \text{diag}[300, 300]$ and $\mathbf{R} = \text{diag}[0.3, 0.3]$ are fixed. We compare the results for two common drag effects: \textit{Linear Drag} from \cite{c5} and \textit{Quadratic Drag}. \textit{Linear Drag:} We use the linear rotor drag model proposed in \cite{c5} where $\mathbf{F}_d(\mathbf{x}) = \mathbf{R}\mathbf{D}\mathbf{R}^T\mathbf{v}$ in (\ref{eq_uav_system}) and $\mathbf{D} = \text{diag}[1, 1, 1]$ is a constant diagonal matrix comprising of the rotor-drag coefficients. In Fig. \ref{fig:lindrag}, our proposed \textit{SOCP Learning} achieves a lower absolute path error over both \textit{SOCP No Learning} and \textit{FMPC} by simultaneously adapting the optimized trajectory and the control input to account for learned GP drag model. Significantly, for faster trajectories $\omega \geq 4.5$, both \textit{FMPC} and \textit{SOCP No Learning} lead to angular constraint violations as a result of neglecting the effects of drag. Consequently, the constrained \textit{SOCP No Learning} optimizer becomes infeasible (marked with a $\times$ in Fig. \ref{fig:lindrag}). \textit{Quadratic Drag:} We use a quadratic drag model, where $\mathbf{F}_d(\mathbf{x}) = \mathbf{R}(\frac{1}{2}\mathbf{\rho}\mathbf{C}_d\mathbf{A}\mathbf{v}_{B}^{2})$, $\rho = 100 $ is the environment density, $\mathbf{C}_d = 0.5$ is the drag coefficient. $\mathbf{A} = 0.159$ is the reference area and $\mathbf{v}_{B}$ is the body frame velocity. As seen in Fig. \ref{fig:squareddrag}, similar to the linear drag case, our proposed approach outperforms related model predictive controllers that neglect the impact of drag, reducing the absolute path error by 15 - 49 \%. In Fig. \ref{fig:squareddrag2}, we see that at lower speeds, $\omega = 2$ rad/s, our proposed \textit{SOCP Learning} achieves this improved tracking by sending larger feasible thrust commands to compensate for drag. In Fig. \ref{fig:squareddrag4.6}, we observe that it is not possible to track the reference given the multirotor constraints. Our proposed approach (\textit{SOCP Learning}) adapts the trajectory to ensure feasible thrust. Neglecting aerodynamic forces (\textit{SOCP No Learning}) leads to constraint violation and infeasibility of the optimal control problem. 

\section{Conclusion}

This paper presents a novel drag-aware model predictive control architecture that optimizes feasible trajectories despite unknown aerodynamic forces. Significantly, our proposed controller can be solved efficiently as a SOCP, facilitating future implementation on resource-constraint multirotor platforms. In future work, we will compare the performance of our proposed approach with GP MPC \cite{c7} in high-speed multirotor experiments. 

%Drag force also impacts constraints on maximum body torques (in terms of dynamically feasible trajectories) which we have neglected. One approach is to similarly learn the mapping from $(z, v)$ to torque as a GP, linearize similar to linGP and ball constraint to approx. this as SOC. 

\addtolength{\textheight}{-0.5cm}


\begin{thebibliography}{99}

\bibitem{c1} E. Kamak, et al., ``Present and future of slam in extreme environments: The darpa subt challenge," \textit{IEEE Transactions on Robotics}, 99:1-20, 2023.

\bibitem{c2} G. Brunner, at al., ``The urban last mile problem: Autonomous drone delivery to your balcony," in \textit{Proc. IEEE International Conference on Unmanned Aircraft Systems (ICUAS)}, 1005–1012, 2019.

\bibitem{c3} L. Eleftherios et al., ``Unsupervised human detection with an embedded vision system on a fully autonomous UAV for search and rescue operations," \textit{Sensors}, 19: ,2019.

\bibitem{c4} R. Mahony, et al., ``Multirotor aerial vehicles: Modeling, estimation, and control of quadrotor," \textit{IEEE Robotics \& Automation Magazine}, 19(3):20–32, 2012.

\bibitem{c5} M. Faessler, et al., ``Differential Flatness of Quadrotor Dynamics Subject to Rotor Drag for Accurate Tracking of High-Speed Trajectories," in \textit{IEEE Robotics and Automation Letters}, 3(2):620-626, 2018.

\bibitem{c6} E. Tal, et al., ``Accurate tracking of aggressive quadrotor trajectories using incremental nonlinear dynamic inversion and differential flatness," \textit{IEEE Transactions on Control Systems Technology}, 29(3):1203–1218, 2020.

\bibitem{c7} G. Torrente, et al., ``Data-driven mpc for quadrotors," \textit{IEEE Robotics and Automation Letters}, 6(2):3769–3776, 2021.

\bibitem{c8} L. Hewing, et al., "Cautious Model Predictive Control Using Gaussian Process Regression," \textit{IEEE Transactions on Control Systems Technology}, 28(6):2736-2743, 2020.

\bibitem{c9} C. J. Ostafew, et al., ``Robust constrained learning-based  NMPC  enabling  reliable  mobile  path-tracking," \textit{International Journal of Robotics Research}, 35(13):1547-1563, 2016.

\bibitem{c10} D. Mellinger et al., ``Minimum snap trajectory generation and
control for quadrotors," in \textit{Proc IEEE Int. Conf. Robot. Autom. (ICRA)}, 2520–2525, 2011.

\bibitem{c11} K. Mohta, et al., ``Fast, autonomous flight in gps-denied and cluttered environments," \textit{Journal of Field Robotics}, 35(1):101–120, 2018.

\bibitem{c12} F. Gao, et al., ``Teach-repeat-replan: A complete and robust system for aggressive flight in complex environments," \textit{IEEE Transactions on Robotics}, 36(5):1526–1545, 2020.

\bibitem{c13} M. Greeff, et al., ``Flatness-based model predictive control for quadrotor trajectory tracking," in \textit{Proc. IEEE/RSJ Int. Conf. on Intelligent Robots and Systems (IROS)}, 6740-6745, 2018.

\bibitem{c14} M. Greeff, et al., ``Exploiting Differential Flatness for Robust Learning-Based Tracking Control Using Gaussian Processes," \textit{IEEE Control Systems Letters} 5(4):1121-1126, 2021.

\bibitem{c15} A. W. Hall, et al., ``Differentially Flat Learning-Based Model Predictive Control Using a Stability, State, and Input Constraining Safety Filter," \textit{IEEE Control Systems Letters}, 7:2191-2196, 2023.

\bibitem{c16} H. Zhang, et al., ``Why Change Your Controller When You Can Change Your Planner: Drag-Aware Trajectory Generation for Quadrotor Systems," ArXiv: https://arxiv.org/abs/2401.04960, 2023. 

\bibitem{c17} C. E. Rasmussen et al., Gaussian processes for machine learning. Cambridge, MA: MIT Press, 2006.

\bibitem{c18} T. X. Nghiem, ``Linearized Gaussian Processes for Fast Data-driven Model Predictive Control," in Proc. \textit{IEEE American Control Conference (ACC)}, 1629-1634, 2019.

\bibitem{c19} M. Diehl, et al., ``Fast direct multiple shooting algorithms for optimal robot control," Springer, 65–93, 2006.

\bibitem{c20} Z. Wu, et al., ``$L_1$Quad: $L_1$ Adaptive Augmentation of Geometric Control for Agile Quadrotors with Performance Guarantees," ArXiv: https://arxiv.org/abs/2302.07208, 2023.

\bibitem{c21} MOSEK Fusion API for C++ 10.1.24.  (2024), https://docs.mosek.com/latest/cxxfusion/index.html

\end{thebibliography}
\end{document}